\pdfoutput=1

\documentclass[11pt]{article}

\usepackage[final]{acl}

\usepackage{times}
\usepackage{latexsym}

\usepackage[T1]{fontenc}

\usepackage[utf8]{inputenc}

\usepackage{microtype}

\usepackage{inconsolata}

\usepackage{graphicx}

\usepackage{multirow}
\usepackage{color}
\usepackage{graphics}
\usepackage{xcolor, soul}
\sethlcolor{lightgray}

\usepackage{booktabs}
\usepackage{amsmath}

\definecolor{aluminum}{RGB}{153,153,153}
\definecolor{platinum}{RGB}{228,228,228}
\definecolor{bgc}{RGB}{245,245,245}
\definecolor{gallery}{RGB}{240,240,240}
\definecolor{tuatara}{RGB}{67, 67, 67}
\definecolor{flamingo}{RGB}{237, 88, 85}
\definecolor{salmon}{RGB}{242,131,107}
\definecolor{free_speech_aquamarine}{RGB}{0, 156, 114}

\definecolor{Aquamarine3}{RGB}{102, 205, 170} 
\definecolor{Goldenrod1}{RGB}{255, 193, 37} 
\definecolor{IndianRed1}{RGB}{255, 106, 106} 
\definecolor{SlateBlue1}{RGB}{131, 111, 255}

\definecolor{bb}{HTML}{95e1d3}
\definecolor{gg}{HTML}{c7ffd8}
\definecolor{yy}{HTML}{f0c38e}
\definecolor{blu}{HTML}{5ab4ba}
\definecolor{rr}{HTML}{f38181}

\definecolor{c1}{HTML}{6E85B2}
\definecolor{c2}{HTML}{368B85}
\definecolor{c3}{HTML}{C56824}
\definecolor{c4}{HTML}{FFC069}
\definecolor{c5}{HTML}{916BBF}

\usepackage{subcaption}

\usepackage{amssymb}
\usepackage{algorithm}
\usepackage{algorithmicx}
\usepackage{algpseudocode}

\usepackage{hyperref}

\usepackage{makecell}

\usepackage[most]{tcolorbox}

\title{SheetDesigner: MLLM-Powered Spreadsheet Layout Generation with Rule-Based and Vision-Based Reflection}

\author{
 \textbf{Qin Chen\thanks{Equal contribution.}\textsuperscript{,1}},
 \textbf{Yuanyi Ren\footnotemark[1]\textsuperscript{,1}},
 \textbf{Xiaojun Ma\thanks{Corresponding author.}\textsuperscript{,2}},
 \textbf{Mugeng Liu\textsuperscript{1}},
 \textbf{Han Shi\textsuperscript{2}},
 \textbf{Dongmei Zhang\textsuperscript{2}},
\\
 \textsuperscript{1}Peking University,
 \textsuperscript{2}Microsoft,
\\
\small \texttt{
\{chenqink,yyren,lmg\}@pku.edu.cn,  \{xiaojunma,shihan,dongmeiz\}@microsoft.com
}
}

\begin{document}
\maketitle


\begin{abstract}
Spreadsheets are critical to data-centric tasks, with rich, structured layouts that enable efficient information transmission. Given the time and expertise required for manual spreadsheet layout design, there is an urgent need for automated solutions.
However, existing automated layout models are ill-suited to spreadsheets, as they often (1) treat components as axis-aligned rectangles with continuous coordinates, overlooking the inherently discrete, grid-based structure of spreadsheets; and (2) neglect interrelated semantics, such as data dependencies and contextual links, unique to spreadsheets. In this paper, we first formalize the spreadsheet layout generation task, supported by a seven-criterion evaluation protocol and a dataset of 3,326 spreadsheets. We then introduce \textbf{SheetDesigner}, a zero-shot and training-free framework using Multimodal Large Language Models (MLLMs) that combines rule and vision reflection for component placement and content population. SheetDesigner outperforms five baselines by at least 22.6\%. We further find that through vision modality, MLLMs handle overlap and balance well but struggle with alignment, necessitates hybrid rule and visual reflection strategies. Our codes and data is available at \href{https://github.com/Cqkkkkkk/SheetDesigner}{Github}. 
\end{abstract}

\section{Introduction}
\label{sec:intro}

Sitting at the heart of finance, analytics, and scientific discovery, spreadsheets serve as powerful tools for organizing and analyzing data \cite{chan1996use,hacker2017financial,powell2019business}. They are structured in a grid of rows and columns with integrated tables and charts. Meanwhile, their effectiveness hinges on clear, well-structured layouts; otherwise, even the most rigorous analysis becomes unreadable when a chart obscures its source data, or a column cuts off text. Consequently, given the importance of well-structured layouts and the time-consuming, expertise-dependent nature of manual design, automated layout generation becomes essential.

However, existing layout generation approaches \cite{zhang2024vascar,kong2022blt,gupta2021layouttransformer,cheng2025graphic} fall short of this task, as they
(1) treat components as rectangles with continuous pixel coordinates, ignoring the grid-based structure of spreadsheets, where components span discrete cells and resizing them affects entire rows or columns.
(2) overlook key semantic relationships, such as placing charts near their source tables, and fail to account for the global row or column resizing to fit content like long text.
Consequently, their outputs must be painstakingly post-processed for spreadsheet layouts and often remain invalid or suboptimal, underscoring a significant and largely unaddressed research problem.

In this paper, we first formalize the task of spreadsheet layout generation: 
Given a raw sheet containing user data (e.g., tables, charts), the goal is to generate a structured layout that enhances spreadsheet usability.
To quantify the evaluation of this task, we introduce an evaluation protocol that scores a candidate layout on seven complementary criteria—\textit{fullness, compactness, compatibility, component-alignment, type-aware alignment, relation-aware alignment,} and \textit{overlap} (see \autoref{sec:preliminary}). We also formulate a dataset, \textit{SheetLayout}, comprising 3,326 real-world spreadsheets covering ten domains and thirteen frequently used topics of functions (see \autoref{table:data-statistics-domain}, \autoref{table:data-statistics-function}).

On this foundation, we propose \textbf{SheetDesigner}, a zero-shot and training-free framework for spreadsheet layout generation, powered by Multimodal Large Language Models (MLLMs). Given a set of user data, SheetDesigner consists of two phases. (1) It initially places components on the grid in a type-aware and relation-aware manner, and subsequently applies a Dual Reflection mechanism, comprising rule-based and vision-based reflection, to refine the layouts. (2) After reflection, it populates the sheet layout with user data, inserts line breaks for lengthy entries, and generates consistent global column widths and row heights, resulting in a ready-to-use layout.

We evaluate SheetDesigner on the SheetLayout dataset against five state-of-the-art baselines, achieving a 22.6\% improvement in performance. Using a 13B-parameter backbone, SheetDesigner matches or exceeds the performance of much larger architectures like LayoutPrompter \cite{lin2023layoutprompter}, leveraging GPT-4O as its backbone \cite{openai2024gpt4ocard}.
Our ablation study highlights the contribution of each component and shows that while the vision modality of MLLMs improves \textit{overlap} and \textit{balance}, they struggle with \textit{alignment}. 
Our further empirical analysis reveals that MLLM attention evidently fails to effectively focus on regions critical for component alignment in structured spreadsheet images, extending the observations of \cite{zhang2025mllms}. 
This highlights the importance of the hybrid rule-based and vision-based reflection mechanisms in SheetDesigner.

In summary, our contributions are as follows:
\begin{itemize}
  \item We present the first task formulation for spreadsheet layout generation, accompanied by a seven-criteria evaluation protocol and a novel dataset, \textit{SheetLayout}, comprising 3,326 spreadsheets spanning 10 common domains and 13 frequent topics.
  \item We introduce \textbf{SheetDesigner}, a zero-shot and training-free framework that directly models spreadsheet layout generation in a two-stage process: \emph{structure placement with Dual Reflection} and \emph{content population with global arrangements}.
  \item We show that SheetDesigner achieves a 22.6\% improvement over five state-of-the-art baselines, with the 13B variant matching or surpassing much larger architectures built on GPT-4o. Our ablation study and empirical analysis further validate the effectiveness of the hybrid rule- and vision-based design.
  

\end{itemize}

\section{Preliminary}
\label{sec:preliminary}


\subsection{Task Formulation}
In this subsection, we formally define the task of spreadsheet layout generation. 
\paragraph{Input} We denote the input raw sheet as $\mathcal{S} = [\mathcal C_1, \mathcal C_2, \dots, \mathcal C_N]$, where each component $\mathcal C_i = \{\mathcal D_i\}$ contains user data $\mathcal D_i$. $\mathcal{D}_i$, including texts, numbers, formulas, etc.

\paragraph{Output} The goal of spreadsheet layout generation is to create a layout: $\mathcal L = [\tilde{\mathcal C}_1, \tilde{\mathcal C}_2, \dots, \tilde{\mathcal C}_N, \mathcal G]$ that organizes the components while preserving their types and formatting the content appropriately.
Each component in the generated layout is represented as $\tilde{\mathcal C}_i = \{\mathcal P_i, \mathcal T_i, \tilde{\mathcal D}_i\}$, where $\mathcal P_i$ specifies the assigned position of the component using the R1C1 format (e.g., "A1:C3"), 
$\mathcal{T}_i$ denotes the assigned type from five component types (e.g., "title"), 
and $\tilde{\mathcal D}_i$ denotes the formatted text with appropriate line breaks. Additionally, $\mathcal G = [w_1, w_2, \dots; h_1, h_2, \dots]$ represents the configuration of the generated layout, encompassing the column widths $w_i$ and row heights $h_i$.

In this paper, we consider the following five common types of spreadsheet components divided by their semantics:  
(1) \textit{title} that provides a descriptive heading for the spreadsheet.  
(2) \textit{main-table} that contains the core structured data, often organized in rows and columns.  
(3) \textit{meta-data} that includes supplementary information such as author names, dates, or version details.  
(4) \textit{summary-data} that presents aggregated insights, such as totals, averages, or key metrics derived from the main table.  
(5) \textit{chart} that visually represents data trends and relationships through graphs, bar charts, or other visual components like images or icons.

\begin{figure*}[t]
    \centering
    \includegraphics[width=\linewidth]{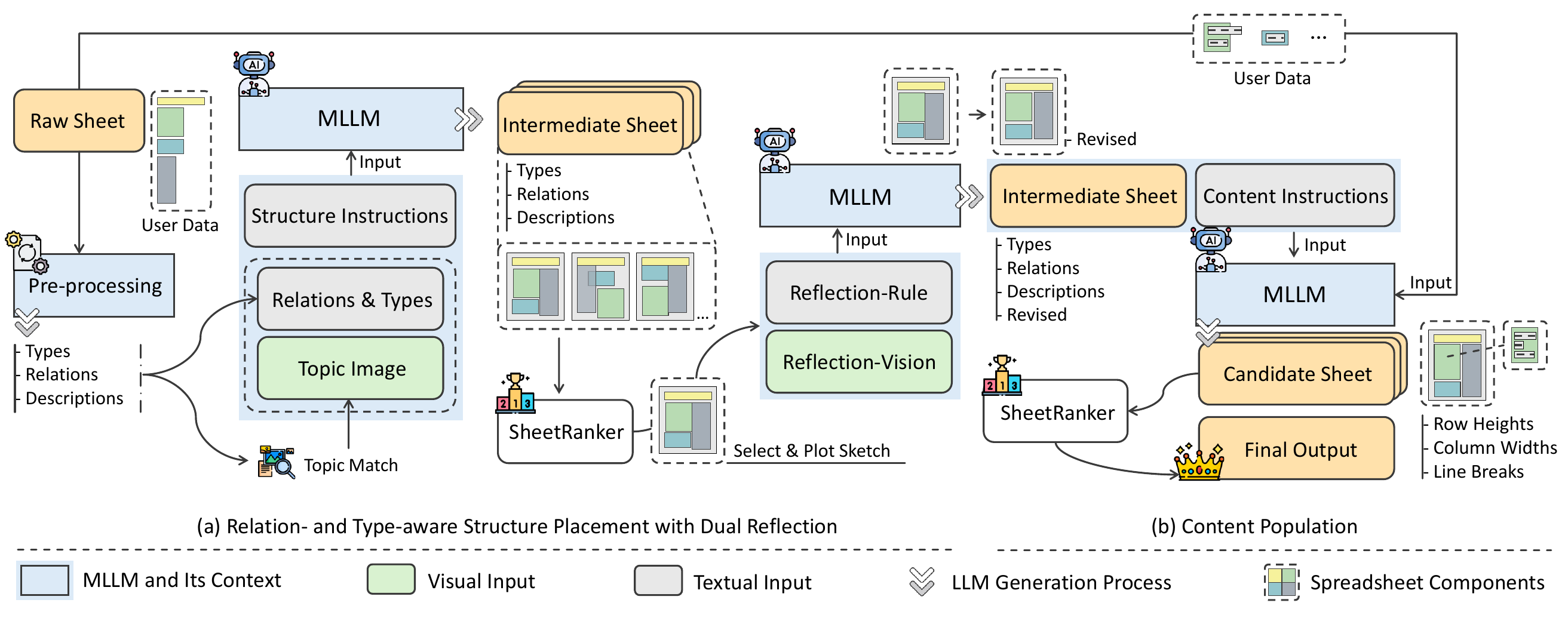}
    \caption{
    SheetDesigner operates in two stages following pre-processing. (a) Components are structurally placed based on their types and relationships. Among multiple layout candidates, SheetRanker selects the best one, which is then refined through Dual-Reflection—a revision step combining rule-based (text) and vision-based (sketch-image) feedback. (b) Content is populated into the placed components, with adjustments to row heights, column widths, and line breaks to ensure proper fit. SheetRanker then selects the final output from the generated candidates.
    }
    \label{fig:framework}
\end{figure*}

\subsection{Layout Evaluation}
\label{sec:eval-protocol}

This subsection outlines the evaluation of generated sheet layouts.
Intuitively, we expect the generated layouts to (1) be compact without large empty space; (2) be well-aligned between components, and the alignment should be broadcast to components of the same \textit{type}, or \textit{dependent} components like table and charts that are drawn by the data in this table ; (3) be visually balanced where there is no great discrepancy for the vertical and horizontal distribution of components; (4) be compatible to the original contents; (5) avoid overlap regions. 
Following this intuition, we define quantitative evaluation metrics (detailed in \autoref{sec:appendix_eval}). 
For \textit{Overlap}, a score of 0 indicates no overlap, with progressively smaller values corresponding to greater overlap.
Other metrics fall within the $(0,1]$ range, where higher scores indicate better performance.

\section{Method}

This section details the structure of SheetDesigner, shown in \autoref{fig:framework}. 
SheetDesigner divides the sheet layout generation into two phases:
(1) \textit{structure placement with Dual Reflection}, which assigns components to appropriate locations by considering both their type and relational context, while ensuring proper alignment.
Then it refines the assignment through rule-based and vision-based reflection; and (2) \textit{content population with global arrangements}, which fills components with user data and sets row heights and column widths based on cell content. The two-phase design leverages MLLMs' strength in handling focused tasks, rather than being distracted by diverse objectives in a single run \cite{wang2024exploring,sun2025fast}.


\paragraph{Pre-processing} Denote each raw sheet as $\mathcal{S} = [\mathcal{C}_1, \mathcal{C}_2, \dots, \mathcal{C}_N]$, where each component $\mathcal{C}_i = \{\mathcal{D}_i\}$ contains the corresponding data $\mathcal{D}_i$. We begin by classifying the entire spreadsheet into one of 13 topics based on its application context, yielding a general topic label $\hat{\mathcal{T}}_{\mathcal{S}}$ for the sheet. Next, for each component $\mathcal{C}_i \in \mathcal{S}$, we:
(\romannumeral 1) assign it a type $\mathcal{T}_i$, selected from five component types (e.g., "title");
(\romannumeral 2) generate a textual description $\hat{\mathcal{D}}_i$ based on its content $\mathcal{D}_i$ (e.g., "A main-table for different services and costs"). We then instruct large language models (LLMs) to identify pairwise relationships between components, resulting in a relation list $\mathcal{R}$ (e.g., $[(\text{Main}_1, \text{Chart}_1)]$). See \autoref{sec:appendix_prompts} for detailed prompt examples. The processed sheet is denoted as $\hat{\mathcal{S}} = \{[\hat{\mathcal{C}}_1, \hat{\mathcal{C}}_2, \dots, \hat{\mathcal{C}}_N], \mathcal{T}_{\mathcal{S}}, \mathcal{R}\}$, where each $\hat{\mathcal{C}}_i = \{\mathcal{T}_i, \hat{\mathcal{D}}_i\}$ includes the assigned type and generated description for the component. We employ LLMs as the pre-processing engine.


\subsection{Structure Placement with Dual Reflection}
\subsubsection{Structure Placement}

In this stage, for each preprocessed sheet $\hat{\mathcal{S}}$, we prompt the MLLMs with: (\romannumeral 1) textual instructions guiding sheet layout generation; (\romannumeral 2) the relations $\mathcal{R}$ and sheet components $[\hat{\mathcal{C}}_1, \hat{\mathcal{C}}_2, \dots, \hat{\mathcal{C}}_N]$, and (\romannumeral 3)  an exemplar topic image $\mathcal{I}$. We generate $N_1$ intermediate sheet layouts per spreadsheet, score them using SheetRanker (see \autoref{sec:sheetranker}), and select the top-performing candidates.

The instructions guide the generation by: (1) preserving alignment among components, particularly in a type-aware manner (aligning components with the same type) and a relation-aware manner (placing related components in proximity); (2) ensuring spatial fullness and balance by maximizing space usage and distributing components evenly in horizontal or vertical; (3) avoiding overlaps. We also provide examples to demonstrate each principle in practice. 
Furthermore, MLLMs are allowed to resize titles and charts to improve layout quality, as these have looser grid constraints than others. Other components are fixed in size. For instance, a 1×5 title can be resized to 1×6 (or 1×4) to enhance alignment without affecting its meaning, whereas resizing a 4×4 data table to 4×5 (or 4×3) would involve adding or removing data. 

For each preprocessed sheet $\hat{\mathcal{S}}$, we retrieve a topic-specific image $\mathcal{I}$, a screenshot of an exemplar spreadsheet of the same type of $\mathcal{T}_{\mathcal{S}}$. We use this image as a visual exemplar to guide the layout of components. Layout conventions, such as spatial grouping, and visual emphasis, are tailored to each document’s topic. For example, recipe cards prioritize vertical lists of ingredients and step-by-step instructions, whereas academic posters allocate prominent space for section headers and data visualizations.

\subsubsection{Dual Reflection} 

Given the top-ranked intermediate sheet layout and its scores for each aspect, we refine it through a dual reflection process:
(1) \textbf{Rule-based Reflection}. For each evaluation aspect, if the SheetRanker score falls below a predefined threshold (e.g., fullness < 0.5), we augment the prompt with targeted revision instructions. For instance, a low overlap score triggers guidance to explicitly avoid component overlap during the reflection step\footnote{If no aspect falls below its threshold, this step is skipped.}.
(2) \textbf{Vision-based Reflection}. We visualize the sheet layout by coloring cells based on component types for an image input. This allows the MLLM to perceive the layout from a visual perspective, enabling further refinement through multimodal understanding.
For detailed information on the threshold for triggering reflection, the specific prompts for revising each aspect, and the algorithm used to generate images of layouts, please refer to \autoref{sec:appendix_method_details}.

\subsection{Content Population and Global Arrangements}
After revising the intermediate sheet, we enter the content-aware stage, where the original data is populated into components with appropriate line breaks. Global sheet layout configurations—specifically, column widths and row heights—are also generated. The prompt includes the following instructions: (1) insert line breaks for lengthy content; (2) adjust column widths and row heights to fit content while minimizing empty space. We also provide examples of common font settings with corresponding row height and column width settings.
The LLM generates $N_2$ candidate sheet layouts, which are then ranked using SheetRanker to produce the final, fully detailed sheet layout.

\subsection{SheetRanker} 
\label{sec:sheetranker} 
Given a set of candidate sheets, SheetRanker assigns a score to each based on the protocol in \autoref{sec:eval-protocol} and selects the one with the highest score. All aspects are weighted equally at 1 \footnote{For aspects with horizontal and vertical sub-aspects, each sub-aspect is weighted at 0.5, maintaining a total weight of 1.}. We adopt this uniform weighting because all aspects (except Overlap) share a common scale of $(0,1]$, while Overlap mostly falls within $(-1,0]$.
SheetRanker serves two key functions:
(\romannumeral 1) guiding selection toward the candidate with the highest overall performance, and
(\romannumeral 2) providing a quantitative foundation for reflecting on and refining structural placement.

\begin{table*}[htbp]
    \small
    \centering
    \caption{
    Quantitative results on SheetLayout reported in mean scores. Scores range from 0 (poor) to 1 (optimal), except for the overlap metric ($\leq0$), where values closer to 0 indicate better performance. The weighted total score assigns a weight of 0.5 to vertical and horizontal sub-aspects, and 1 to all other aspects. Relative performance is reported with respect to the best-performing model. For LLM-based methods, the underlying language model is indicated in parentheses. "C" denotes Component, "T" for Type-aware, and "R" for Relation-aware.
    }
    \label{table:main}
    \resizebox{.999\linewidth}{!}{
    \setlength{\tabcolsep}{3pt} 
    \begin{tabular}{lcccccccccccccccccc}
        \toprule
         &  \multirow{2}{*}{\textbf{Fullness}} & \multicolumn{2}{c}{\textbf{Compatibility}} & \multicolumn{2}{c}{\textbf{C-Alignment}} & \multicolumn{2}{c}{\textbf{T-Alignment}} & \multicolumn{2}{c}{\textbf{R-Alignment}} & \multicolumn{2}{c}{\textbf{Balance}} & \multirow{2}{*}{\textbf{Overlap}} & \multirow{2}{*}{\makecell[c]{\textbf{Weighted Total}}} \\
         \cmidrule(l{2pt}r{2pt}){3-4} \cmidrule(l{2pt}r{2pt}){5-6}\cmidrule(l{2pt}r{2pt}){7-8} \cmidrule(l{2pt}r{2pt}){9-10}\cmidrule(l{2pt}r{2pt}){11-12}
         &  & Horizontal & Vertical & Horizontal & Vertical & Horizontal & Vertical & Horizontal & Vertical & Horizontal & Vertical & \\ 
         \midrule

BLT & 0.485 & 0.285 & 0.586 & 0.373 & 0.604 & 0.379 & 0.571 & 0.451 & 0.547 & 0.474 & 0.504 & -0.184 & 2.688 ($\downarrow$ 45.12\%)\\
LayoutFormer++ & 0.618 & 0.308 & 0.559 & 0.501 & 0.728 & 0.407 & 0.585 & 0.556 & 0.607 & 0.595 & 0.705 & -0.125 & 3.268 ($\downarrow$ 33.27\%)\\
Coarse-to-Fine & 0.531 & 0.289 & 0.576 & 0.475 & 0.635 & 0.402 & 0.601 & 0.519 & 0.528 & 0.601 & 0.725 & -0.143 & 3.063 ($\downarrow$ 37.45\%)\\
\midrule
PosterLLaVa (LLaVA-7B) & 0.653 & 0.376 & 0.608 & 0.404 & 0.712 & 0.430 & 0.642 & 0.609 & 0.684 & 0.610 & 0.732 & -0.183 & 3.373 ($\downarrow$ 31.12\%)\\
LayoutPrompter (GPT-4o) & 0.804 & 0.397 & 0.623 & 0.508 & 0.789 & 0.487 & 0.683 & 0.690 & 0.716 & 0.634 & 0.778 & -0.167 & 3.789 ($\downarrow$ 22.63\%)\\
\midrule
SheetDesigner (Vicuna-7B) & 0.703 & 0.395 & 0.617 & 0.434 & 0.778 & 0.485 & 0.628 & 0.662 & 0.581 & 0.545 & 0.683 & -0.103 & 3.504 ($\downarrow$ 28.46\%)\\
SheetDesigner (LLaVA-7B) & 0.706 & 0.431 & 0.629 & 0.458 & 0.794 & 0.486 & 0.637 & 0.690 & 0.585 & 0.619 & 0.721 & -0.075 & 3.656 ($\downarrow$ 25.36\%)\\
SheetDesigner (Vicuna-13B) & 0.678 & 0.424 & 0.649 & 0.456 & 0.803 & 0.521 & 0.678 & 0.675 & 0.642 & 0.668 & 0.753 & -0.056 & 3.756 ($\downarrow$ 23.31\%)\\
SheetDesigner (LLaVA-13B) & 0.690 & 0.432 & 0.661 & 0.459 & 0.806 & 0.530 & 0.680 & 0.695 & 0.668 & 0.696 & 0.793 & -0.043 & 3.857 ($\downarrow$ 21.25\%)\\
SheetDesigner (GPT-4o) & \hl{0.981} & \hl{0.549} & \hl{0.886} & \hl{0.683} & \hl{0.880} & \hl{0.788} & \hl{0.858} & \hl{0.703} & \hl{0.679} & \hl{0.894} & \hl{0.920} & \hl{-0.003} & \hl{4.898}\\

        \bottomrule
        \end{tabular}
        }
\end{table*}

\begin{table*}[!htbp]
    \small
    \centering
    \caption{Ablation study, "w/o" denotes "without".}
    \label{table:ablation-100percel-full}
    \resizebox{.999\linewidth}{!}{
    \setlength{\tabcolsep}{3pt} 
    \begin{tabular}{lcccccccccccccccccc}
        \toprule
         &  \multirow{2}{*}{\textbf{Fullness}} & \multicolumn{2}{c}{\textbf{Compatibility}} & \multicolumn{2}{c}{\textbf{C-Alignment}} & \multicolumn{2}{c}{\textbf{T-Alignment}} & \multicolumn{2}{c}{\textbf{R-Alignment}} & \multicolumn{2}{c}{\textbf{Balance}} & \multirow{2}{*}{\textbf{Overlap}} & \multirow{2}{*}{\makecell[c]{\textbf{Weighted Total}}} \\
         \cmidrule(l{2pt}r{2pt}){3-4} \cmidrule(l{2pt}r{2pt}){5-6}\cmidrule(l{2pt}r{2pt}){7-8} \cmidrule(l{2pt}r{2pt}){9-10}\cmidrule(l{2pt}r{2pt}){11-12}
         &  & Horizontal & Vertical & Horizontal & Vertical & Horizontal & Vertical & Horizontal & Vertical & Horizontal & Vertical &  \\ 
         \midrule

SheetDesigner & 0.981 & 0.549 & 0.886 & 0.683 & 0.880 & 0.788 & 0.858 & 0.703 & 0.679 & 0.894 & 0.920 & -0.003 & 4.898\\
\midrule
w/o Topic & 0.973 & 0.537 & 0.859 & 0.653 & 0.856 & 0.736 & 0.821 & 0.683 & 0.673 & 0.876 & 0.897 & -0.003 & 4.766 ($\downarrow$ 2.71\%)\\
w/o Reflection-Rule & 0.956 & 0.530 & 0.876 & 0.641 & 0.820 & 0.744 & 0.793 & 0.642 & 0.664 & 0.823 & 0.856 & -0.007 & 4.644 ($\downarrow$ 5.20\%)\\
w/o Reflection-Vision & 0.941 & 0.544 & 0.879 & 0.672 & 0.867 & 0.788 & 0.854 & 0.698 & 0.674 & 0.852 & 0.882 & -0.012 & 4.784 ($\downarrow$ 2.33\%)\\
w/o Reflection & 0.925 & 0.520 & 0.853 & 0.622 & 0.805 & 0.739 & 0.781 & 0.628 & 0.631 & 0.804 & 0.802 & -0.017 & 4.500 ($\downarrow$ 8.12\%)\\
w/o SheetRanker & 0.916 & 0.491 & 0.752 & 0.637 & 0.842 & 0.729 & 0.784 & 0.673 & 0.659 & 0.859 & 0.884 & -0.010 & 4.561 ($\downarrow$ 6.88\%)\\
w/o Vision & 0.926 & 0.522 & 0.701 & 0.638 & 0.848 & 0.705 & 0.804 & 0.632 & 0.632 & 0.823 & 0.874 & -0.015 & 4.500 ($\downarrow$ 8.12\%)\\

        \bottomrule
        \end{tabular}
        }
\end{table*}

\section{Experiments}
In this section, we conduct experiments to verify the effectiveness of the proposed SheetDesigner. 

\subsection{Dataset}

For the evaluation of our model, we construct a dataset, \textit{SheetLayout}, consisting of 3,326 Excel spreadsheets collected from various domains and real-world applications (See \autoref{table:data-statistics-domain} and \autoref{table:data-statistics-function}. The dataset encompasses diverse spreadsheet structures, reflecting practical use cases across multiple fields. We perform object detection within each spreadsheet, identifying key components. These detected objects are subsequently converted into a structured JSON format to enable standardized processing. See \autoref{sec:appendix_dataset} for details on the collection, anonymous process, and licenses.

\subsection{Baselines \& Settings}
We compare the proposed \textit{SheetDesigner} with various state-of-the-art baselines, which can be broadly categorized into two groups:  
\begin{itemize}
    \item \textbf{Traditional Transformer-based models}, trained and validated on layout datasets, including BLT \cite{kong2022blt}, LayoutFormer++ \cite{jiang2023layoutformer}, and Coarse-to-Fine \cite{jiang2022coarse}.
    \item \textbf{LLM-based approaches}, which leverage LLMs to enable few-shot or zero-shot layout generation, including LayoutPrompter \cite{lin2023layoutprompter} and PosterLLaVA \cite{yang2024posterllava}. For LayoutPrompter we adopt GPT-4o as the backbone, as the recommended \texttt{text-davinci-003} is deprecated. 
\end{itemize}

While the aforementioned baselines perform well in general layout generation, they are neither specifically designed nor optimized for spreadsheet layouts. Their outputs are typically pixel-based bounding boxes, like $[(x_1, y_1, x_2, y_2), \dots]$, where each box defines the top-left and bottom-right pixel coordinates of a component. 
To adapt these layouts for spreadsheets, we introduce a standardized procedure that maps pixel-based layouts to a grid-based structure. We assume a $\mathcal B_x\times \mathcal B_y$ pixel background, where each grid cell corresponds to a $\mathcal C_x\times \mathcal C_y$ pixel area, yielding a $\frac{\mathcal B_x}{\mathcal C_x}\times \frac{\mathcal B_y}{\mathcal C_y}$ grid. If any layout exceeds $\mathcal B_x$ pixels in width (or $\mathcal B_y$ pixels in height), we scale it down proportionally to fit within these constraints. Components that do not align perfectly with the grid (i.e., whose positions are not exact multiples of $\mathcal C_x$ or $\mathcal C_y$ pixels) are adjusted by snapping to the nearest cell. 

In this study, we adopt $\mathcal{B}_x = 1000$, $\mathcal{C}_x = 50$, $\mathcal{B}_y = 500$, and $\mathcal{C}_y = 25$, reflecting typical settings in commonly used spreadsheet applications. The threshold for triggering Dual Reflection is set to a moderate value of $0.5$ for all aspects except Overlap, which uses a strict threshold of $0$, as any intersection significantly degrades layout usability and therefore always triggers a revision. The number of repeated runs is set to $N_1 = N_2 = 3$. To ensure a fair comparison between training-based and training-free methods, the data is split into 10\% for training, 10\% for validation, and 80\% for testing. All reported performance metrics are based on the test set.
We adopt three families of backbone models: GPT-4o \cite{openai2024gpt4ocard}, the Vicuna family \cite{zheng2023judging}, and the LLaVA family \cite{liu2023visual}. Regarding modality, GPT-4o and LLaVA are multimodal LLM with vision ability, while Vicuna models are text-only. For Vicuna, vision-related inputs are disabled. The LLaVA models utilized in this study employ their corresponding Vicuna models as the backbone LLM.

\subsection{Evaluation Results}

In \autoref{table:main}, we compare SheetDesigner with several baselines on the dataset SheetLayout. 
Employing GPT-4o as its backbone, SheetDesigner demonstrates state-of-the-art performance. It surpasses the second-place LayoutPrompter, which also utilizes GPT-4o, by a notable 22.63\% in total score.
Variants using smaller models also remain competitive. Notably, when equipped with a 13B model (either Vicuna-13B or LLaVA-13B), SheetDesigner performs competitively or even surpasses LayoutPrompter—which relies on the much stronger GPT-4o backbone—demonstrating the effectiveness of our framework. Additionally, at comparable parameter scales, models with vision perception (LLaVA family) consistently outperform those without (Vicuna family). 

\begin{figure*}[htbp]
    \centering
    \includegraphics[width=\linewidth]{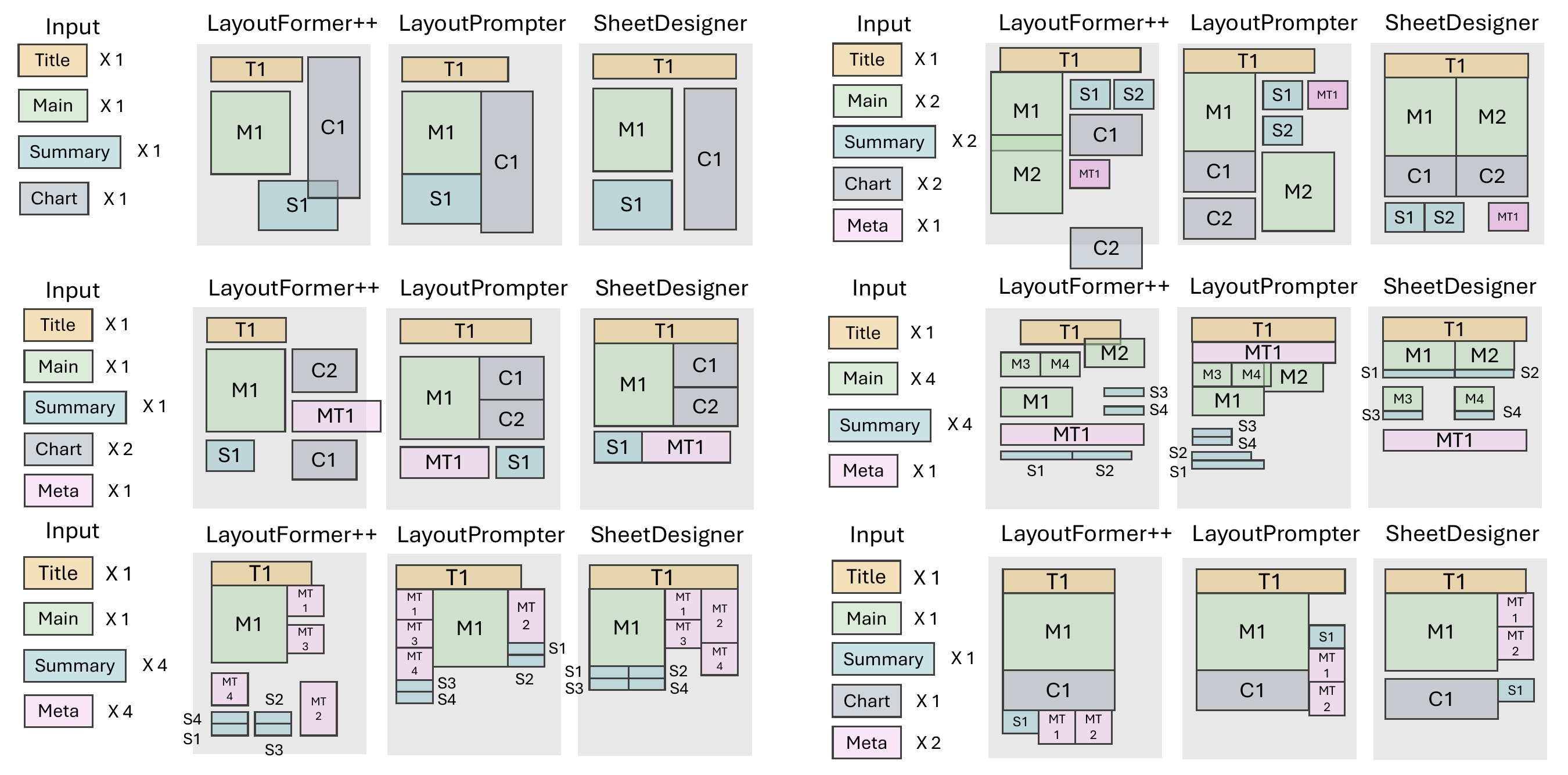}
    \caption{Qualitative comparison among SheetDesigner, LayoutFormer++ and LayoutPrompter. We denote each component with the letters, \texttt{T} for titles, \texttt{M} for main tables, \texttt{S} for summary tables, and \texttt{MT} for metadata tables.}
    \label{fig:case_good}
\end{figure*}

We visualize some of the layouts generated by SheetDesigner, LayoutFormer++, and LayoutPrompter for comparison in \autoref{fig:case_good}. 
Generally SheetDesigner achieves favorable results against the baselines. Compared to LayoutFormer++ and LayoutPrompter, SheetDesigner produces more aligned layouts. Further, SheetDesigner arranges components in a relation-aware and type-aware manner, for example, in \autoref{fig:case_good} (middle-right), SheetDesigner places summary tables of corresponding tables directly below them, assuring better readability and usefulness compared to the others.
For a analysis on failure cases, please refer to \autoref{sec:appendix_case_study}.


\subsection{Ablation Study}
\label{sec:experiment_ablation_study}

\begin{figure*}[htbp]
    \centering
     \includegraphics[width=\linewidth]{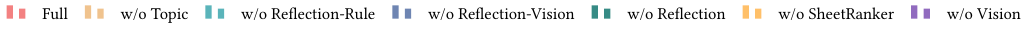}
     \includegraphics[width=0.285\linewidth]{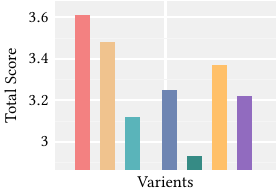}
    \includegraphics[width=0.7\linewidth]{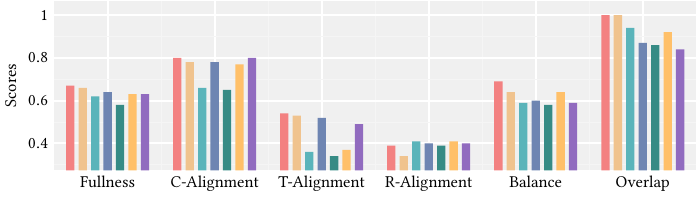}
    
   \caption{Ablation study for the lower 1\% tail of the score distributions. An offset of 1 was added to the overlap scores for clarity. Aspects with horizontal and vertical sub-components were merged by averaging their scores.}
    \label{fig:ablation-1percent}
\end{figure*}

To assess the contribution of each component in SheetDesigner, we conduct an ablation study (see \autoref{table:ablation-100percel-full}), including a “w/o Vision” case in which both the topic image and vision-based reflection are removed, and visualize the lowest 1\% of the score distributions (see \autoref{fig:ablation-1percent}). Our results show that every component enhances performance to varying degrees: the SheetRanker yields a significant overall improvement; the topic images notably boosts balance and relation-aware alignment; rule-based reflection affects almost every metric (its smallest impact being on overlap); and vision-based reflection improves balance and reduces overlap but contributes little to alignment measures. Generally, the visual modality of MLLMs is helpful for improving balance or overlap, but struggle with alignment requirements.


\begin{table}[h]
    \centering
    \caption{Quantitative comparison: Specified dimensions with line wraps than the AutoFit of Excel. Relative performance degradation is reported.}
    \label{table:specified-vs-autofit}
    \resizebox{.999\linewidth}{!}{
    \begin{tabular}{lcccccccccccccccccc}
        \toprule
          &  \multicolumn{2}{c}{\textbf{Compatibility}} & \multirow{2}{*}{\makecell[c]{\textbf{Deg. Ratio}}} \\
          \cmidrule{2-3}
          &  Horizontal & Vertical & \\
                  
         \midrule
Specified (Full) & 0.549 & 0.886 & - \\
AutoFit (Full) & 0.538 & 0.853 & 3.07\% \\
\midrule
Specified (\%5 Low) & 0.523 & 0.882 & - \\
AutoFit (\%5 Low) & 0.495 & 0.824 & 6.12\% \\
\midrule
Specified (\%1 Low) & 0.521 & 0.896 & - \\
AutoFit (\%1 Low) & 0.485 & 0.818 & 8.05\% \\

        \bottomrule
        \end{tabular}
        }
\end{table}
We compared our proposed ContentPopulator with Excel’s built-in AutoFit function (see \autoref{table:specified-vs-autofit}) across the full dataset, as well as the lowest 5\% and 1\% of scores. The results support our claim that explicitly specifying row and column dimensions, along with appropriate line wrapping, generally outperforms AutoFit.

\subsection{Hyper-parameter Analysis on Thresholds in Dual Reflection}
\label{sec:hyper_dual_reflection_thredhold}

\begin{table}[htbp]
    \centering
    \caption{Hyper-parameter Analysis on the thresholds in Dual Reflection. We report the total scores, and the average tokens costs with GPT-4o as the backbone.}
    \label{table:hyper-parameter}
    \setlength{\tabcolsep}{3pt}
    \resizebox{.999\linewidth}{!}{
    \begin{tabular}{lcccccccccccccccccc}
        \toprule
         & Total Score $\uparrow$ & Dual Reflection Token Cost $\downarrow$\\             
         \midrule
        Threshold=0.3 & 4.604 &  103.5 \\
        Threshold=0.5 & 4.898 &  234.2 \\
        Threshold=0.7 & 4.904 &  616.7 \\
        \bottomrule
        \end{tabular}
        }
\end{table}

The thresholds in Dual Reflection determine when a layout needs revision. We use a moderate threshold of 0.5 for the main experiments, as stated previously. Lower thresholds (e.g., 0.3) reduce computation by targeting only extreme cases, while higher thresholds (e.g., 0.7) improve quality but increase computation due to more frequent revisions. In this section, we show results for thresholds 0.3 and 0.7 in \autoref{table:hyper-parameter}. A 0.3 threshold significantly reduces performance, while 0.7 offers little improvement but uses many more tokens. This indicates that some layouts need revision, but not all are fixable, for example, some spreadsheets are inherently difficult due to mixed object sizes. This supports our choice of a moderate threshold for efficient performance gains.

\begin{figure*}[htbp]
    \centering
    \includegraphics[width=0.325\linewidth]{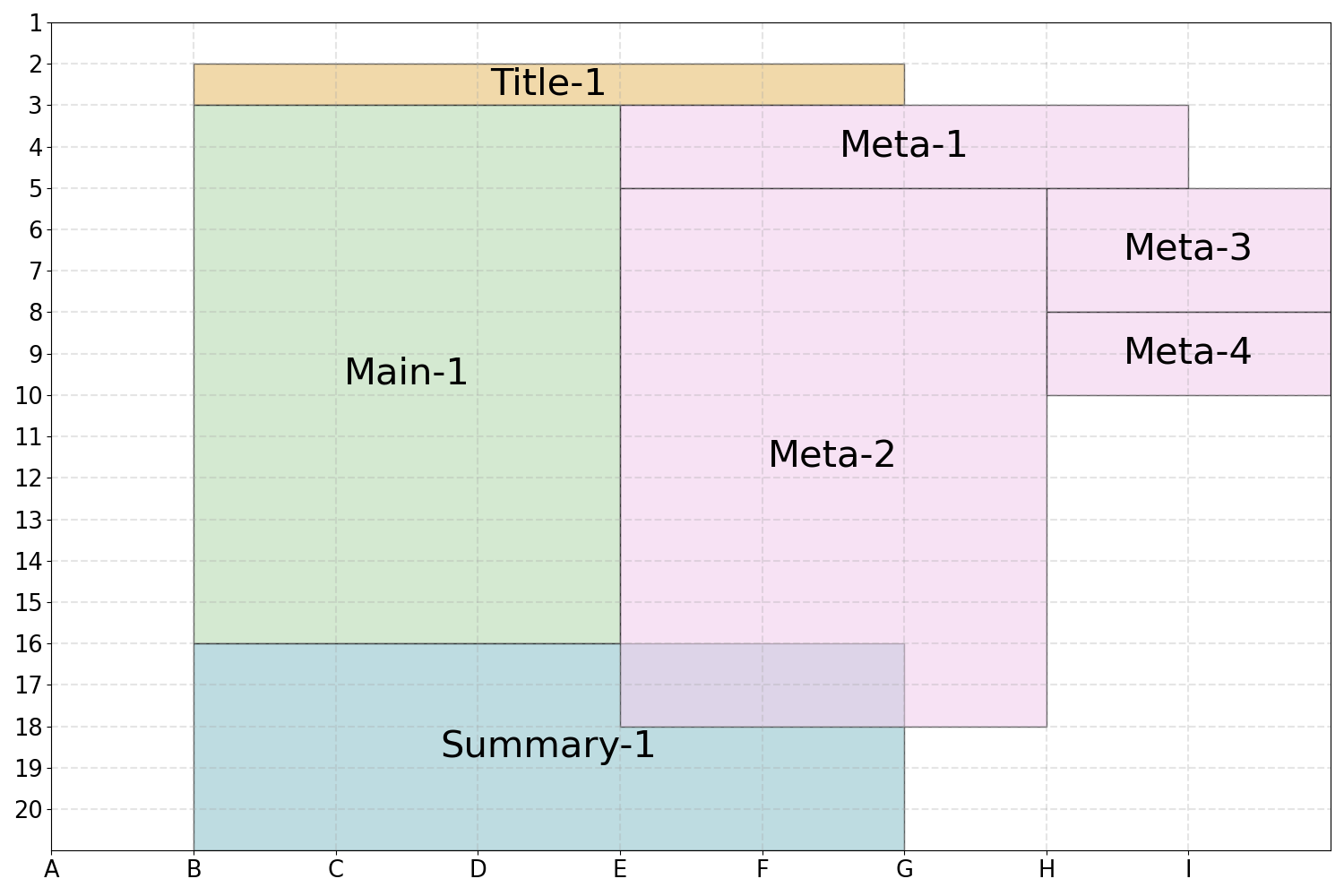}
    \includegraphics[width=0.325\linewidth]{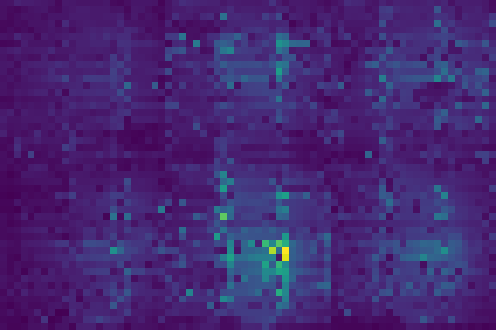}
    \includegraphics[width=0.325\linewidth]{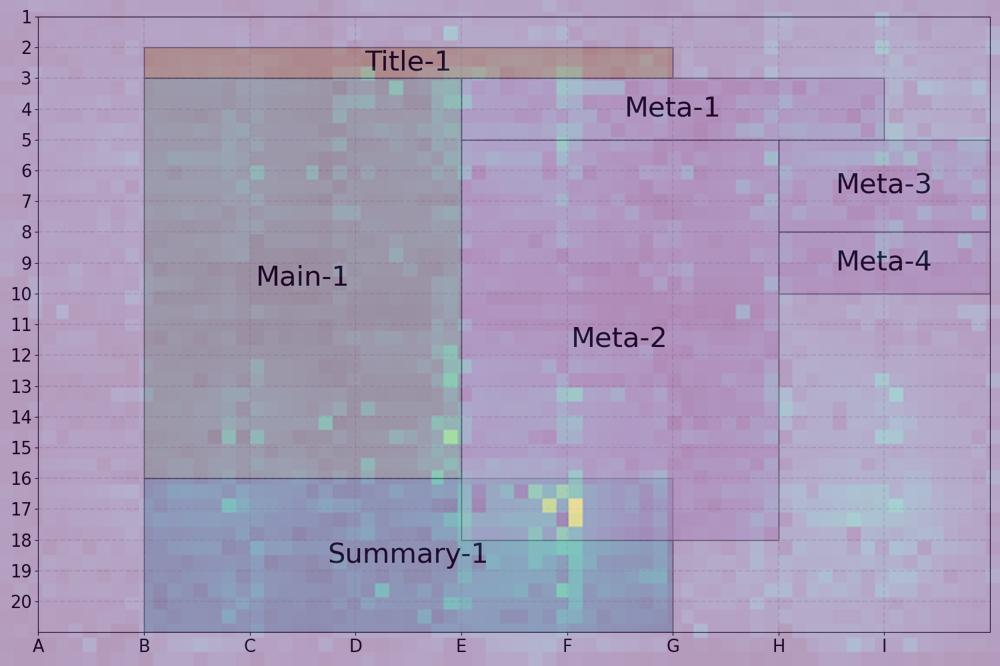}
    \includegraphics[width=0.325\linewidth]{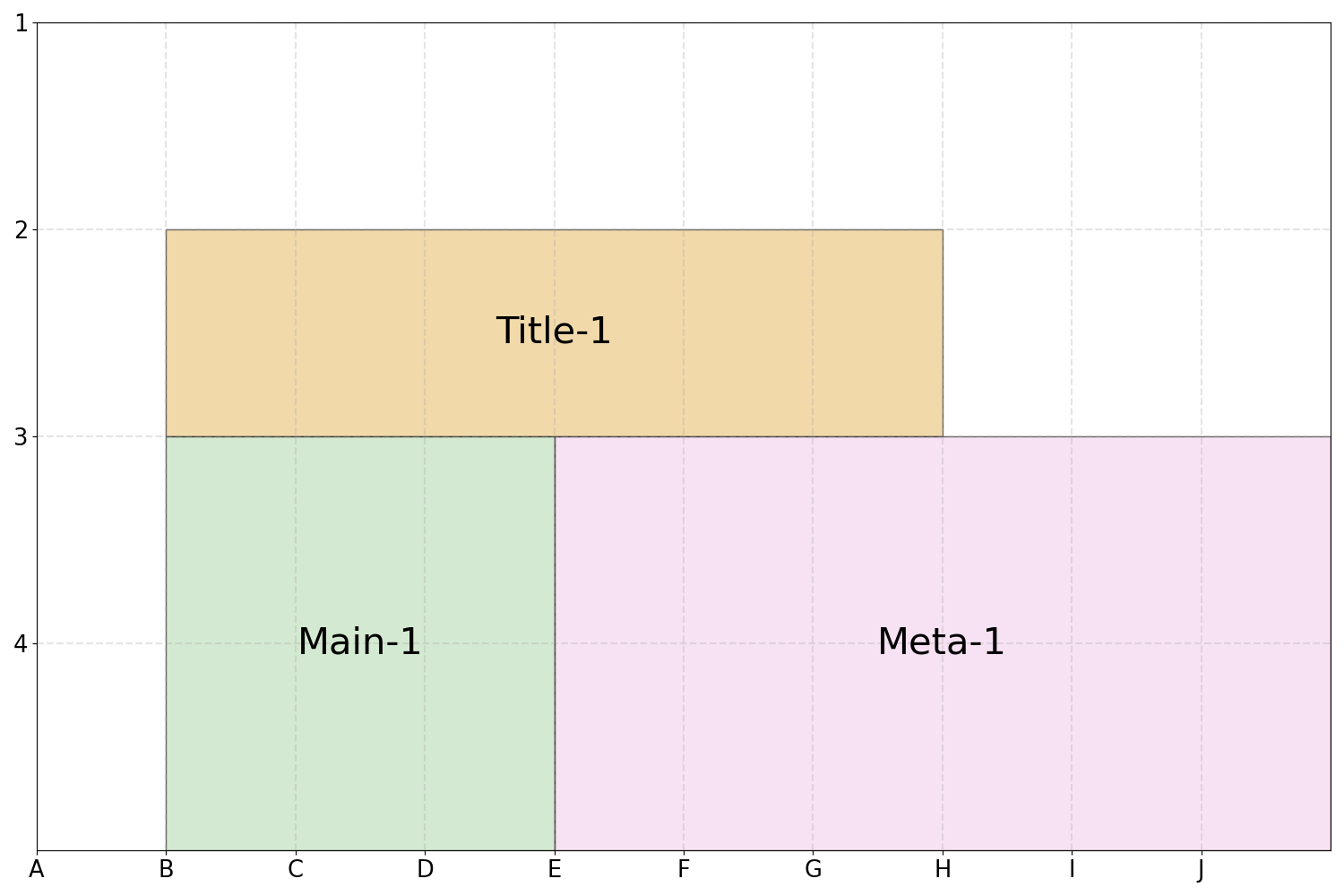}
    \includegraphics[width=0.325\linewidth]{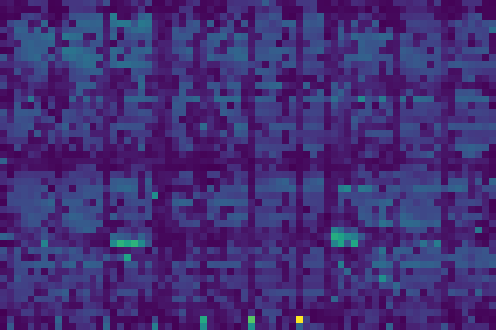}
    \includegraphics[width=0.325\linewidth]{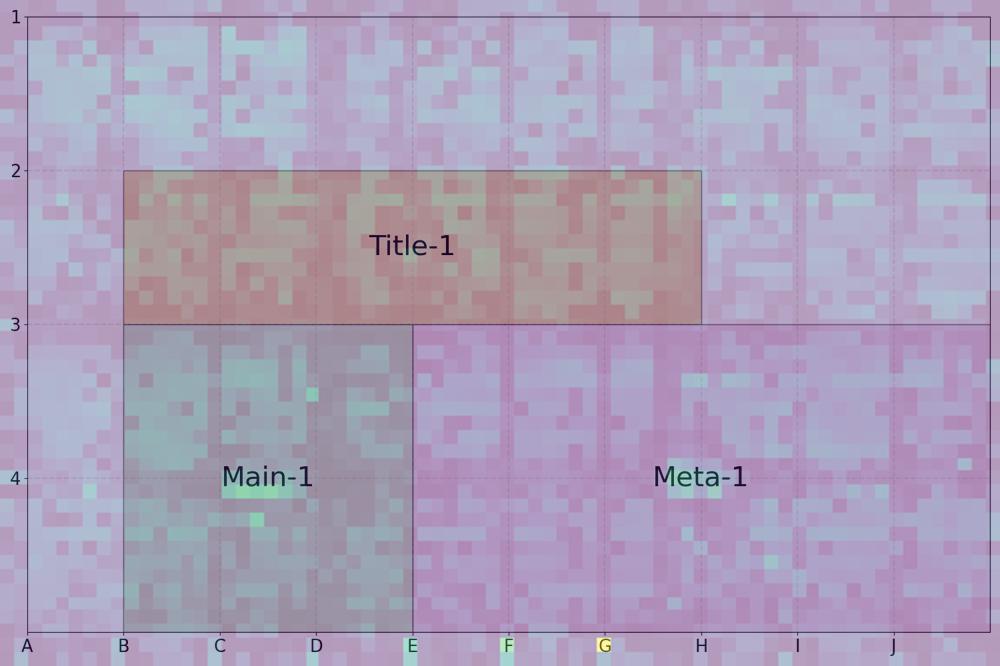}
   \caption{
   Visualization of attention weights on the input image. The first row shows an instance with overlapping components, where LLaVA-7B demonstrates precise attention with a concentrated weight distribution. In contrast, the second row illustrates an instance with misalignment, where the model's attention is scattered and fails to accurately capture misaligned components, such as the right border of the title block.
   }
    \label{fig:attention-weights}
\end{figure*}

\section{Why Does Vision Help Balance and Overlap but Not Alignment?}

As previously stated in \autoref{sec:experiment_ablation_study}, our findings indicate that with visual input, MLLMs excel at improving overlap, balance but struggle with alignment-related features. To further investigate this, we visualize the attention weights of LLaVA-7B using two sketch images, following the setup in \cite{zhang2025mllms}. One layout contains overlapping components, while the other exhibits misaligned elements. In each case, the model is prompted to identify regions of either overlap or misalignment. To facilitate direct visual interpretation of the attention weights, we use the general prompt \textit{"detect the spreadsheet’s components"} to normalize the attention maps \cite{DBLP:conf/acl/LiuHYY25}.

As shown in \autoref{fig:attention-weights}, LLaVA-7B demonstrates precise attention to overlapping regions, indicated by a highly concentrated distribution of weights. Conversely, its attention is scattered and disorganized when dealing with misalignment, failing to accurately identify misaligned components like the right border of the title block. 
The model's proficiency with overlap, akin to identifying a "man with a yellow backpack," likely stems from its optimization for perceiving natural objects with mixed visual features. However, alignment demands a fine-grained focus on the boundaries between paired components, an area where MLLMs currently lack sufficient optimization. This disparity in processing may explain the limited contribution of vision to alignment in \autoref{fig:ablation-1percent}.


This analysis underscores the pivotal role of our Dual Reflection module, which leverages the complementary strengths of rule-based (textual) and vision-based (image) reasoning. In alignment tasks, textual reasoning demonstrates a clear advantage due to the explicitness of coordinate data. For instance, the alignment between "A1:A3" and "B1:B3," contrasted with the misalignment with "C2:C4," can be directly inferred from positional information. However, recognizing this relationship in images requires detailed pairwise visual reasoning. Conversely, for spatial features such as overlap or balance, visual perception provides immediate and intuitive insights, while textual analysis demands additional processing to extract the same information from raw coordinates. The Dual Reflection module’s strength lies in integrating these two modalities, resulting in notable performance gains. These findings also highlight opportunities for improving MLLMs, particularly by enhancing their visual reasoning capabilities \cite{wang2024exploring}—a critical need when interpreting structured formats like spreadsheet images.

\section{Related Works}
\paragraph{Traditional Pixel-oriented Layout Generation}
Layout generation is a widely studied topic encompassing various sub-tasks, including: (1) generation conditioned on element types \cite{kikuchi2021constrained,kong2022blt,lee2020neural,arroyo2021variational}, (2) generation conditioned on both element types and sizes \cite{kong2022blt,cheng2024colay}, (3) generation conditioned on element relationships \cite{kikuchi2021constrained,lee2020neural,cheng2024colay}, (4) layout completion \cite{gupta2021layouttransformer}, (5) layout refinement \cite{rahman2021ruite}, (6) content-aware generation \cite{hsu2023posterlayout,zheng2019content,zhang2024vascar}, (7) text-to-layout generation \cite{huang2021creating,lin2023parse}, (8) layout revision \cite{li2024revision}, and more. Transformers and diffusion models are popular and powerful backbones for these tasks.

\paragraph{LLM-driven Layout Generation} Beyond the traditional methods mentioned above, recent studies \cite{lin2023layoutprompter,yang2024posterllava,tang2023layoutnuwa,seol2024posterllama,zhang2025smaller,hsu2025postero,tang2024layoutkag} have explored leveraging Large Language Models (LLMs) for layout generation. These approaches offer benefits such as zero-shot capability, robustness, multi-task generalization, and strong generation performance, all without the need for task-specific training. Additionally, efforts are being made to enhance LLMs with vision modalities \cite{cheng2025graphic}, Chain-of-thought reasoning \cite{shi2025layoutcot}, and diffusion models \cite{liu2025efficient}. 


While considerable research has been conducted in general layout generation, the resulting methods may not directly align with the distinct characteristics of spreadsheets. Spreadsheets inherently demand strict conformity to a grid, contrasting with approaches that permit arbitrary pixel-level element positioning. Moreover, element dimensions within spreadsheets, such as row heights and column widths, function as global parameters; an adjustment to column \texttt{A}'s width, for example, consequently alters all cells in that column. Finally, spreadsheet layouts inherently require awareness of component types, relationships, and content, which many general-purpose layout approaches do not fully incorporate.

\section{Conclusion}
In this paper, we formalize the task of spreadsheet layout generation, develop an evaluation protocol covering seven key aspects, and present a dataset comprising 3,326 spreadsheets.
We then introduce SheetDesigner for this task, a zero-shot, and training-free framework driven by multimodal large language models. SheetDesigner adopts a two-stage strategy. SheetDesigner involves (1) structural placement with Dual Reflection and (2) content population with global arrangements. Experimental results reveal SheetDesigner's superior performance, surpassing baselines by a significant 22.63\% in performance. Ablation studies reveal the impact of each component. We further conduct a further empirical analysis of MLLMs' vision capabilities. This analysis underscores the necessity of our hybrid Dual Reflection module. The study also illuminate key considerations for advancing MLLMs in the future.

\section{Limitations}
This work has the following limitations:
(i) The dataset is limited to a set of commonly encountered fields, potentially missing the unique requirements and challenges of less-represented or novel domains. This may impact the generalizability of our findings and highlights the need for future work to incorporate more diverse field types to more comprehensively evaluate SheetDesigner's performance.
(ii) As shown in \autoref{sec:appendix_case_study}, the model arranges elements sub-optimally in extreme cases involving a large number of components with varying sizes. This limitation is not unique to our approach and has also been reported in prior work \cite{arroyo2021variational,lin2023layoutprompter}. Addressing such cases remains an open direction for future research.
\bibliography{sample-base}

\appendix
\section{Potential Risks}
This work focuses on automating the generation of spreadsheet layouts through SheetDesigner. While automation offers clear efficiency and usability benefits, it also introduces certain risks. In particular, over-reliance on automatically generated layouts may lead to reduced user control or unintended formatting choices that do not align with domain-specific expectations. Additionally, if deployed in high-stakes environments (e.g., finance or healthcare), errors in layout generation could affect data interpretation and decision-making \cite{chen2025large}. It is therefore important to incorporate mechanisms for human oversight, validation, and customization to mitigate these risks and ensure responsible deployment.

\section{Declaration of Generative AI Tools}

Generative AI tools were used solely for the purpose of language polishing. All ideas, experiments, analyses, and writing were originally developed by the authors.

\section{Evaluation of Spreadsheet Layouts}
\label{sec:appendix_eval}
In this section we detail the seven aspects of evaluation metrics. All metrics, except for \textit{Overlap}, are scaled from $(0,1]$, where higher scores indicate better performance. For \textit{Overlap}, a score of 0 signifies no overlaps, while increasingly negative scores reflect a greater degree of overlap.

\paragraph{\textbf{Fullness}} This aspect evaluates the spatial utilization of generated layouts \cite{hsu2023posterlayout}. We first identify the top-left and bottom-right corners to determine the background region size \( R_\text{bg} \). Next, we mark the areas occupied by layout components, denoted as \( R_\text{ft} \). Note that when calculating area sizes, the row and column dimensions \( G \) are taken into account.
The \textit{fullness} metric is defined as:

\begin{equation}
\small
    S_\text{full}([\tilde C], G) =  
    \begin{cases}
        1, & \text{if } \frac{\text{size}(R_\text{ft})}{\text{size}(R_\text{bg})} \geq \theta_\text{full}, \\
        \frac{\text{size}(R_\text{ft})}{\text{size}(R_\text{bg})}, & \text{otherwise}.
    \end{cases}
\end{equation}
where \(\text{size}(\cdot)\) represents the two-dimensional area measurement, considering both row and column dimensions, $\theta_\text{full}$ is a threshold value. This metric encourages the generation of compact and practically useful spreadsheet layouts while allowing sufficient space for line breaks and separation where necessary by assigning full scores for fullness greater than $\theta_\text{full}$. 

\paragraph{\textbf{Compatibility}} This aspect evaluates how well the provided row heights and column widths accommodate the corresponding text within each cell \cite{hsu2023posterlayout}. To achieve this, we first approximate the average pixel dimensions of the text, assuming a width of $\mathcal W_\text{text}$ pixels per character and a height of $\mathcal H_\text{text}$ pixels per line. We then convert the row heights and column widths into pixel units via . The compatibility scores in both the horizontal and vertical directions are defined as the normalized average compatibility scores across all cells:
\begin{equation}
    \begin{aligned}
            S_\text{compt\_h} & = \frac{1}{1 + \frac{1}{M}|\sum_{i=1}^{M} \frac{S_hw_i}{\mathcal W_\text{text}l_i + P_h} - 1|}, \\
            S_\text{compt\_v} & =  \frac{1}{1 + \frac{1}{M}|\sum_{i=1}^{M} \frac{h_i}{\mathcal H_i n_i + P_v} - 1|}.\\
    \end{aligned}
\end{equation}
Here, $M$ represents the total number of data-containing cells. The variables $w_i$ and $h_i$ denote the width and height of the $i$-th cell, respectively. The term $l_i$ corresponds to the number of characters in the cell's content, while $n_i$ represents the number of text lines in the cell. $S_h$ denotes the factor translating spreadsheet cell width to pixels. $P_h, P_v$ denotes the padding space for horizontal and vertical.
We apply the shifted reciprocal transformation $f(x)=\frac{1}{1+|x-1|}$ to normalize the scores. This ensures that cells that are either too wide or too narrow for the given text receive lower scores, while optimal compatibility results in a score approaching 1.

\paragraph{\textbf{Component Alignment}} Alignment lies in the core of assessing a layout's practical usage and beauty \cite{li2020attribute}. We begin by measuring general alignment between components and then extend the evaluation to type-aware and relation-aware alignment. Given a list of components, we identify frequently occurring positions and assess alignment. Deviations from these positions contribute to an alignment violation score $S_\text{vio\_h}$ and $S_\text{vio\_v}$. Formally, we detect the top-$k$ most frequent positions in both directions and check whether each component aligns with them. A perfect match results in no violation; otherwise, the violation score ($S_\text{vio\_h}$ or $S_\text{vio\_v}$) increases by one. We then normalize the final alignment scores via the reciprocal transformation $f(x)=\frac{1}{1+x}$, ensuring that perfect alignment results in a score of 1, while greater misalignment lowers the score:
\begin{equation}
    \begin{aligned}
    \small
        S_\text{align\_h}  = \frac{1}{1 + \frac{1}{N}  S_\text{vio\_h}}, \;
        S_\text{align\_v} = \frac{1}{1 + \frac{1}{N}  S_\text{vio\_v}}.\\
    \end{aligned}
\end{equation}

\paragraph{\textbf{Type-aware Alignment}} Type-aware alignment focus on measuring how components of the same type aligns with each other. We classify the components by their type, and calculate the alignment scores within each group. The final score of type-aware alignment is calculated by averaging the scores between the types.

\paragraph{\textbf{Relation-aware Alignment}} Relation-aware alignment evaluates how referenced components correspond to each other. For instance, in a main table and its summary table, proper alignment ensures visual hierarchy and perceptual integrity. We detect the groups of related components and measure the alignment within each group. The final score of relation-aware alignment is calculated by averaging the scores between the groups.

\paragraph{\textbf{Balance}} Balance assesses whether components are evenly distributed to maintain visual equilibrium in a spreadsheet \cite{lavie2004assessing}. A well-balanced layout should be achieved both vertically and horizontally, avoiding excessive weight on one side, such as clustering components on the left while leaving the right sparsely populated. Technically, the spreadsheet is divided vertically and hierarchically into two parts\footnote{Components spanning the midpoint are proportionally allocated to both parts}. The \textit{fullness} metric is then applied to each part, and the final scores are computed as:
\begin{equation}
    \small
    \begin{aligned}
        S_\text{balance\_h} &= 1 - \frac{\left|S_\text{full}([\tilde C]_\text{left},G)-S_\text{full}([\tilde C]_\text{right},G)\right|}{S_\text{full}([\tilde C]_\text{left},G)+S_\text{full}([\tilde C]_\text{right},G)} \\
         S_\text{balance\_v} &= 1 - \frac{\left|S_\text{full}([\tilde C]_\text{upper},G)-S_\text{full}([\tilde C]_\text{down},G)\right|}{S_\text{full}([\tilde C]_\text{upper},G)+S_\text{full}([\tilde C]_\text{down},G)},
    \end{aligned}
\end{equation}
where \( [\tilde{C}]_{\text{name}} \) represents the list of components belonging to the corresponding part. For a well-balanced layout, the balance score is 1. The more imbalanced the layout, the lower the score.

\paragraph{\textbf{Overlap}} 
Overlap assesses whether components occupy previously assigned areas \cite{li2020attribute}. We iterate through all components, marking the background to compute the overlap score, where each pair of collision increases the overlap count \( C_\text{overlap} \) by 2. The final score is given by:
\begin{equation}
    S_\text{overlap} = - \frac{C_\text{overlap}^2}{N}
\end{equation}
We impose a strong penalty on overlap by applying a quadratic term, $C_\text{overlap}^2$, since even minor overlaps severely undermine practical utility. A perfectly non-overlapping layout receives a score of 0, while overlapping components are penalized with increasingly negative values.  
Theoretically, the scores fall within the range $[-N(N-1)^2, 0]$, where every pair of components overlaps. In practice, however, the scores typically lie within $(-1, 0)$ in most cases.

\paragraph{Settings} We set $\theta_\text{full}=0.8$ for \textit{fullness}. For \textit{compatibility}, we approximate the $\mathcal W_\text{text}=12$ and $\mathcal H_\text{text}=15$ for the default settings of Calibri with a font size of 12 in English in this paper. For other fonts and languages, the corresponding constants can be adjusted accordingly. We set the padding terms $P_h=40$ and $P_v=10$. The translating factor $S_h$ is set to 7.

We do not use metrics like Fréchet Inception Distance to assess similarity to real layouts. While real layouts can serve as useful references, they are not the only valid designs. Effective spreadsheet layouts can be arranged in many different ways. There is no definitive ground truth for what a layout should be. Instead, we evaluate generated layouts using the seven criteria described above. A layout does not need to resemble existing ones to be useful. If it scores well on these criteria, it can still be effective for practical applications. This evaluation approach also encourages diversity in the generated layouts.

\section{Additional Methodology Details}
\label{sec:appendix_method_details}

\subsection{Exemplar Image $\mathcal I_{\mathcal S}$}

\begin{figure*}[htbp]
    \centering
    \includegraphics[width=\linewidth]{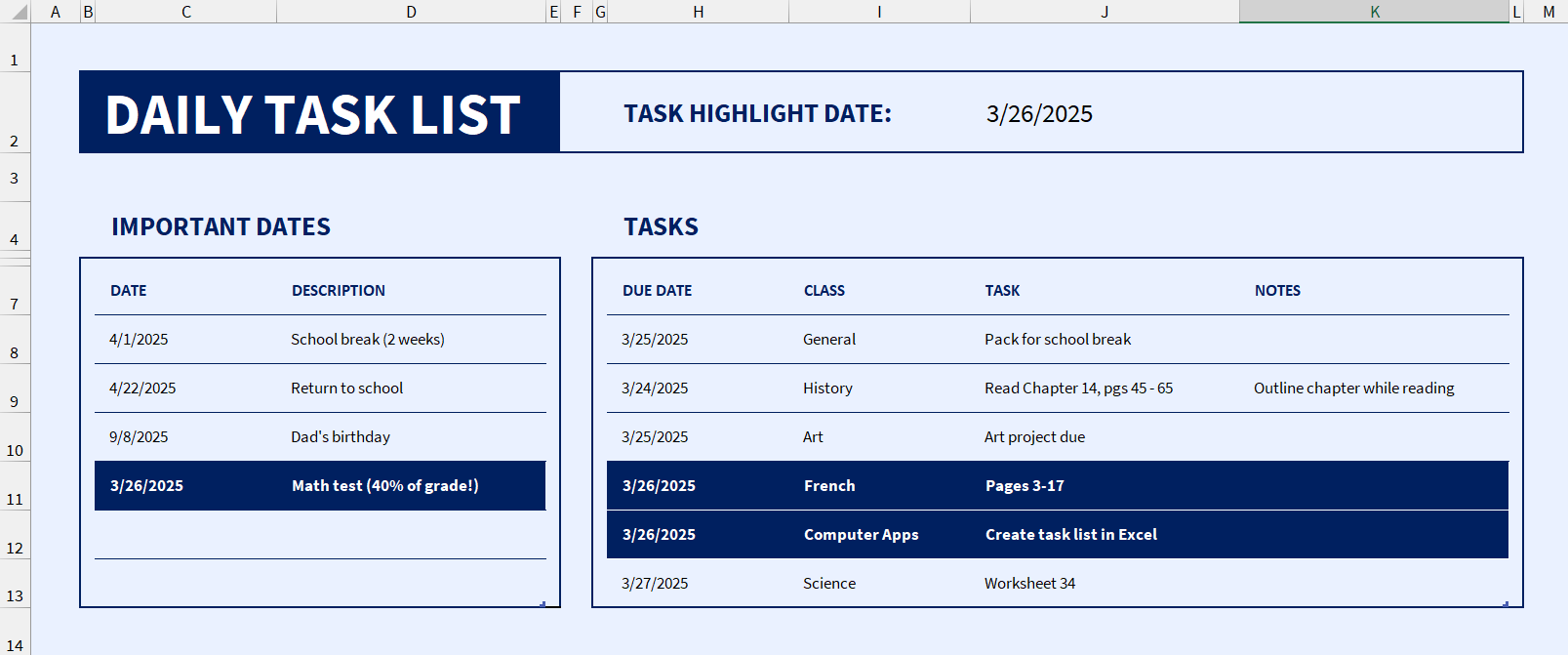}
    \caption{An exemplar image of topic "To-do Lists and Calendars".}
    \label{fig:topic-example}
\end{figure*}

During structural placement, we provide an exemplar image $\mathcal{I}_{\mathcal{S}}$ to the MLLMs as a reference for topic-aware component arrangement. This exemplar is selected based on the topic or application context, as the topic significantly influences layout structures. For example, given the topic "Check List," layouts from domains such as finance, education, IT, or healthcare tend to share common structural patterns. We provide an example in \autoref{fig:topic-example}

In our implementation, we curate 5–10 exemplar images for each of the 13 general spreadsheet topics in \autoref{table:data-statistics-function}. For each structural placement task, one exemplar from the same topic is randomly selected. Importantly, all exemplars are excluded from the SheetLayout test set, ensuring there is no risk of information leakage.

\subsection{Sketch Image Generation for Layouts}

\begin{figure*}[htbp]
    \centering
    \includegraphics[width=0.495\linewidth]{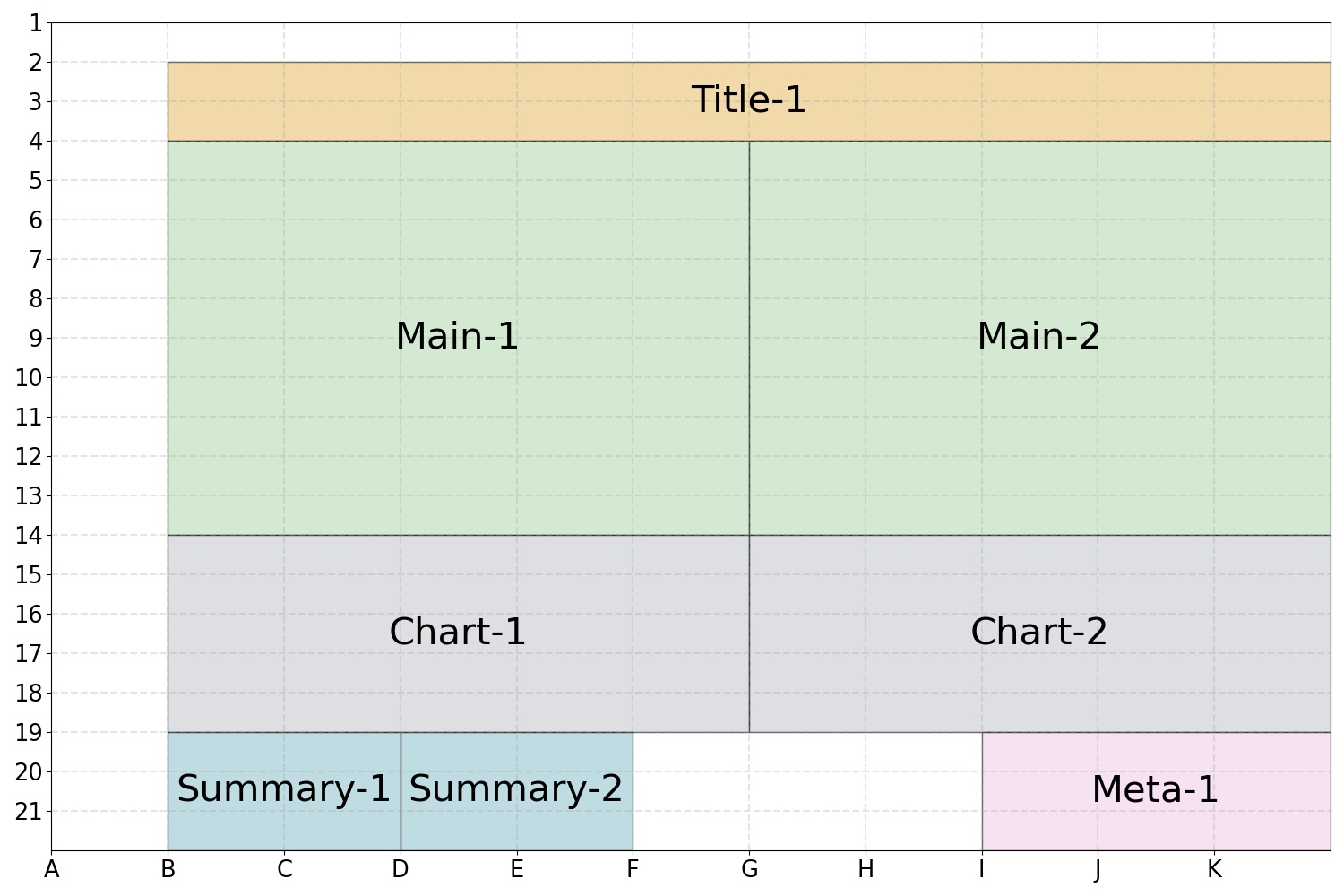}
    \includegraphics[width=0.495\linewidth]{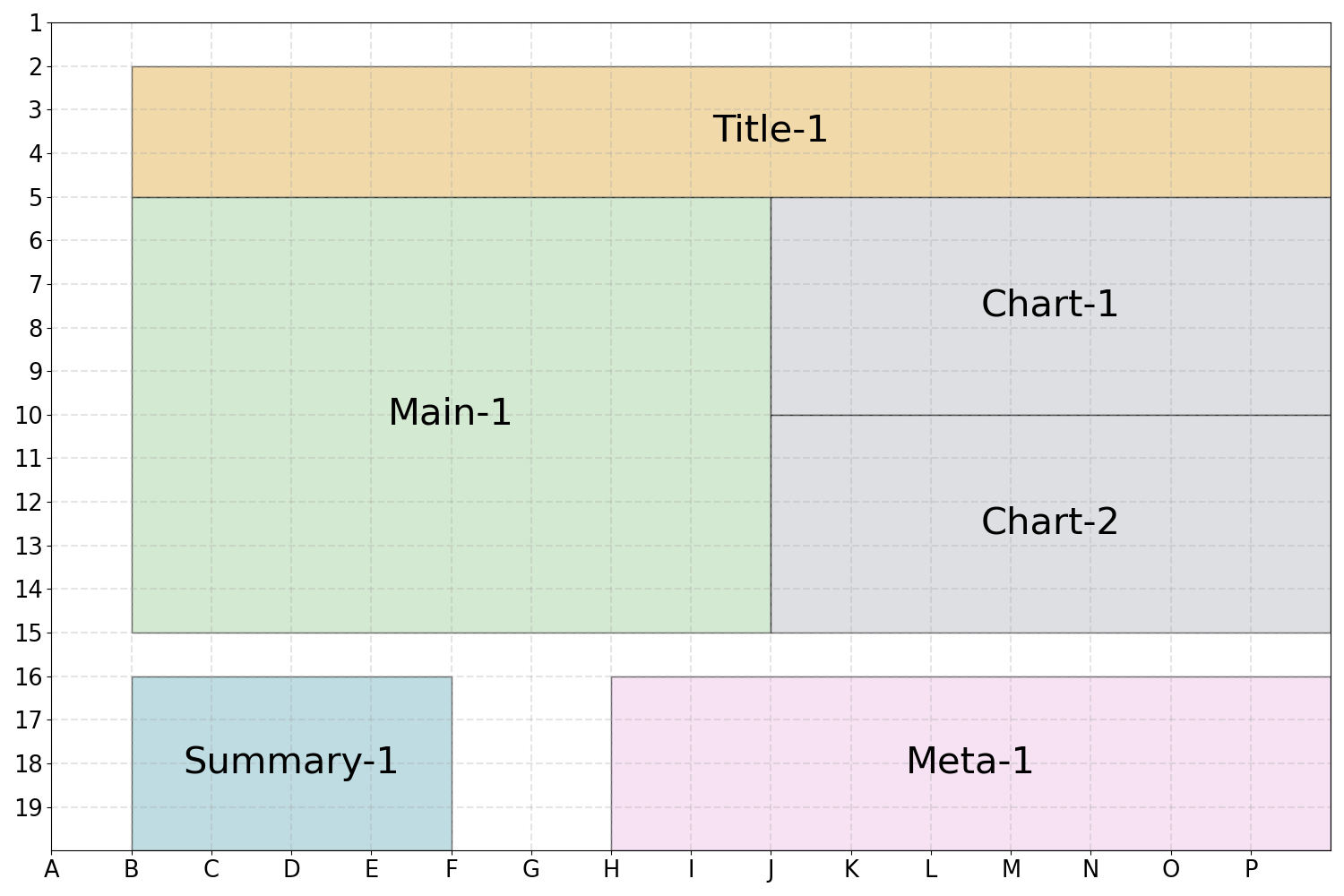}
    \caption{Examples of the sketch images generated from layouts.}
    \label{fig:sketch-image-example}
\end{figure*}

For the sketch image of generated layouts, we first detect the maximum grid size of the layout and define the background. Then, for each component we color the corresponding cells with corresponding texts. Different types of elements are colored with different colors. This sketch image provide clear visual information of the fullness, alignment (and its variants), balance, and overlap. We provide the algorithm in \autoref{algo:plot_sketch_image}, and an example of the images in \autoref{fig:sketch-image-example}.

\begin{algorithm}[H]
\caption{PlotComponent}
\begin{algorithmic}[1]
\Require $K$ component, $\mathcal{C}$ canvas, $\mathcal{S}$ style map
\State $L \gets K.\mathrm{layout}.\mathrm{location}$
\State $(c_{s},e_{c}) \gets \mathrm{PhraseLocation}(L)$
\State $(r_{s},c_{s}) \gets \mathrm{CellToIndex}(c_{s})$
\State $(r_{e},c_{e}) \gets \mathrm{CellToIndex}(e_{c})$
\State $\Delta c \gets c_{e} - c_{s} + 1$
\State $\Delta r \gets r_{e} - r_{s} + 1$
\State $\mathit{fill}\gets \mathcal{S}[\,K.\mathrm{type}\,]$
\State $\mathrm{DrawRectangle}(\mathcal{C},c_{s},r_{s},\Delta c,\Delta r,\mathit{fill})$
\State $\mathrm{DrawText}(\mathcal{C},K.\mathrm{id},c_{s}+\tfrac{\Delta c}{2},r_{s}+\tfrac{\Delta r}{2})$
\end{algorithmic}
\end{algorithm}

\begin{algorithm}[H]
\caption{PlotLayout}
\label{algo:plot_sketch_image}
\begin{algorithmic}[1]
\Require $\Lambda$ layout data
\State $\mathcal{C}\gets \mathrm{InitCanvas}()$
\State $\mathcal{S}\gets \mathrm{DefineStyles}()$
\State $(R_{\max},C_{\max})\gets \mathrm{ComputeGridSize}(\Lambda)$
\State $\mathrm{ConfigureCanvas}(\mathcal{C},R_{\max},C_{\max})$
\ForAll{$\mathcal{K}\in\Lambda$}
    \State \Call{PlotComponent}{$K,\mathcal{C},\mathcal{S}$}
\EndFor
\end{algorithmic}
\end{algorithm}

\subsection{Thresholds and Instructions in Dual Reflection}


We present the reflection-triggering thresholds in \autoref{table:dual-reflection-specific-thresholds}. If any aspect score from SheetRanker falls below its corresponding threshold, the relevant instruction from \autoref{table:dual-reflection-specific-instructions} is appended to the reflection prompt. All scores, except \textit{Overlap}, are on a unified scale of $(0,1)$; thus, we adopt a moderate threshold of $0.5$. For \textit{Overlap}, where any intersection degrades layout usability, a strict threshold of $0$ is used—any overlap triggers revision.

These thresholds are tunable hyper-parameters. Lower values (e.g., 0.3) reduce computation by filtering only extreme cases, while higher values (e.g., 0.7) enhance quality at the cost of increased processing due to more frequent revisions.

\begin{table}[htbp]
    \centering
    \caption{Thresholds for triggering specific instructions in Dual Reflection.}
    \label{table:dual-reflection-specific-thresholds}
    \begin{tabular}{lll}
        \toprule
        Aspect & Threshold \\
        \midrule
        Fullness & 0.5 \\
        Overlap & 0.0 \\
        Alignment & 0.5 \\
        T-Alignment & 0.5 \\
        R-Alignment & 0.5 \\
        Balance & 0.5 \\
        \bottomrule
        \end{tabular}
\end{table}

\begin{table*}[htbp]
    \small
    \centering
    \caption{Specific instructions in Dual Reflection.}
    \label{table:dual-reflection-specific-instructions}
    \resizebox{.999\linewidth}{!}{
    \begin{tabular}{lll}
        \toprule
        Aspect & Instruction \\
        \midrule
        Fullness & This spreadsheet is with much empty space. Consider redistribute the elements to minimize empty space. \\
        Overlap & This spreadsheet has overlapping components. Consider moving the components to avoid overlapping \\
        Alignment & The horizontal alignment of components is not good. Consider align the top of the components \\
        Alignment & The vertical alignment of components is not good. Consider align the left of the components \\
        T-Alignment & The type-specific horizontal alignment of components is not good. Consider align the top of the components according to their types \\
        T-Alignment & The type-specific vertical alignment of components is not good. Consider align the left of the components according to their types \\
        R-Alignment & The relation-specific horizontal alignment of components is not good. Consider align the top of the components according to their relations \\
        R-Alignment & The relation-specific vertical alignment of components is not good. Consider align the left of the components according to their relations \\
        Balance & The horizontal balance of components is not good. Consider distribute the components horizontally \\
        Balance & The vertical balance of components is not good. Consider distribute the components vertically \\
        \bottomrule
        \end{tabular}
        }
\end{table*}

\section{Dataset Details}
\label{sec:appendix_dataset}

In this section, we summarize the dataset statistics. The 3,326 spreadsheets span 10 domains (see \autoref{table:data-statistics-domain}) and cover 13 common topics grouped by the application context (see \autoref{table:data-statistics-function}).

Datasets were acquired from a variety of public platforms. These include official government open data websites (e.g., data.gov) that share public reports or tools in spreadsheet formats; repositories for open science data, such as Zenodo \footnote{https://zenodo.org}, where researchers deposit supplementary materials; and digital libraries like the Internet Archive, which preserve publicly accessible documents. All pre-existing cell content within these publicly sourced spreadsheets was subsequently anonymized using offline LLMs, which replaced original data with synthetically generated content that is semantically coherent with the original data's type and structure. The specific terms and numerical values were excluded from the input and the output cell content was generated entirely by privacy-secure LLMs, ensuring no sensitive data was retained. This process preserves the structural and contextual integrity of the dataset while mitigating privacy risks. 

For every acquired spreadsheet, its specific licensing and terms of use were meticulously verified to ensure suitability for academic research and inclusion in this layout-focused dataset. Preference was given to content in the public domain or under permissive open licenses, such as Creative Commons (e.g., CC0, CC BY).

In addition to these publicly sourced materials, the dataset incorporates spreadsheet layouts originating from Microsoft 365 Create \footnote{https://create.microsoft.com/en-us} published online. The incorporation of these anonymized layouts for academic research was authorized under a formal agreement with Microsoft. This agreement required the thorough anonymization of all original cell content prior to inclusion in the dataset.

\begin{table}[H]
    \centering
    \caption{Statistics of the dataset on domain distribution.}
    \label{table:data-statistics-domain}
        \begin{tabular}{lccccccc}
            \toprule
            Domain & \#Sheets \\
            \midrule
            Business and Finance  & 445 \\
            Marketing and Sales & 397 \\
            Engineering and Manufacturing & 386 \\
            Sports and Entertainment & 320 \\
            Healthcare and Medical & 330 \\
            Education and Research & 355\\
            Personal  and Daily Life & 302\\
            Technology and IT & 290 \\
            Agriculture and Food & 260\\
            Hospitality and Tourism & 241 \\
            \midrule
            \textbf{Total} & 3326 \\
            \bottomrule
        \end{tabular}
\end{table}

\begin{table}[H]
    \centering
    \caption{Statistics of the dataset on topic distribution.}
    \label{table:data-statistics-function}
    \resizebox{.999\linewidth}{!}{
        \begin{tabular}{lccccccc}
            \toprule
            Topic & \#Sheets \\
            \midrule
            Financial Management and Forecasting  & 499 \\
            Data and Task Logs & 394 \\
            Staff Scheduling and Shift Management & 301 \\
            Performance and KPI Dashboards  & 298 \\
            Event Scheduling and Planning & 288 \\
            Inventory and Asset Management  & 269\\
            Report and Publication Tracking & 267 \\
            Maintenance Scheduling  & 155 \\
            Marketing Campaign Tracking & 151 \\
            Project Scheduling & 150 \\
            To-do Lists and Calendars & 180 \\
            Travel Itinerary and Planning  & 165 \\
            Goal and Habit Tracking  & 159 \\
            \midrule
            \textbf{Total} & 3326 \\
            \bottomrule
        \end{tabular}
    }
\end{table}

\section{Case Study}
\label{sec:appendix_case_study}

\subsection{Failure Cases}

\begin{figure}[htbp]
    \centering
    \includegraphics[width=\linewidth]{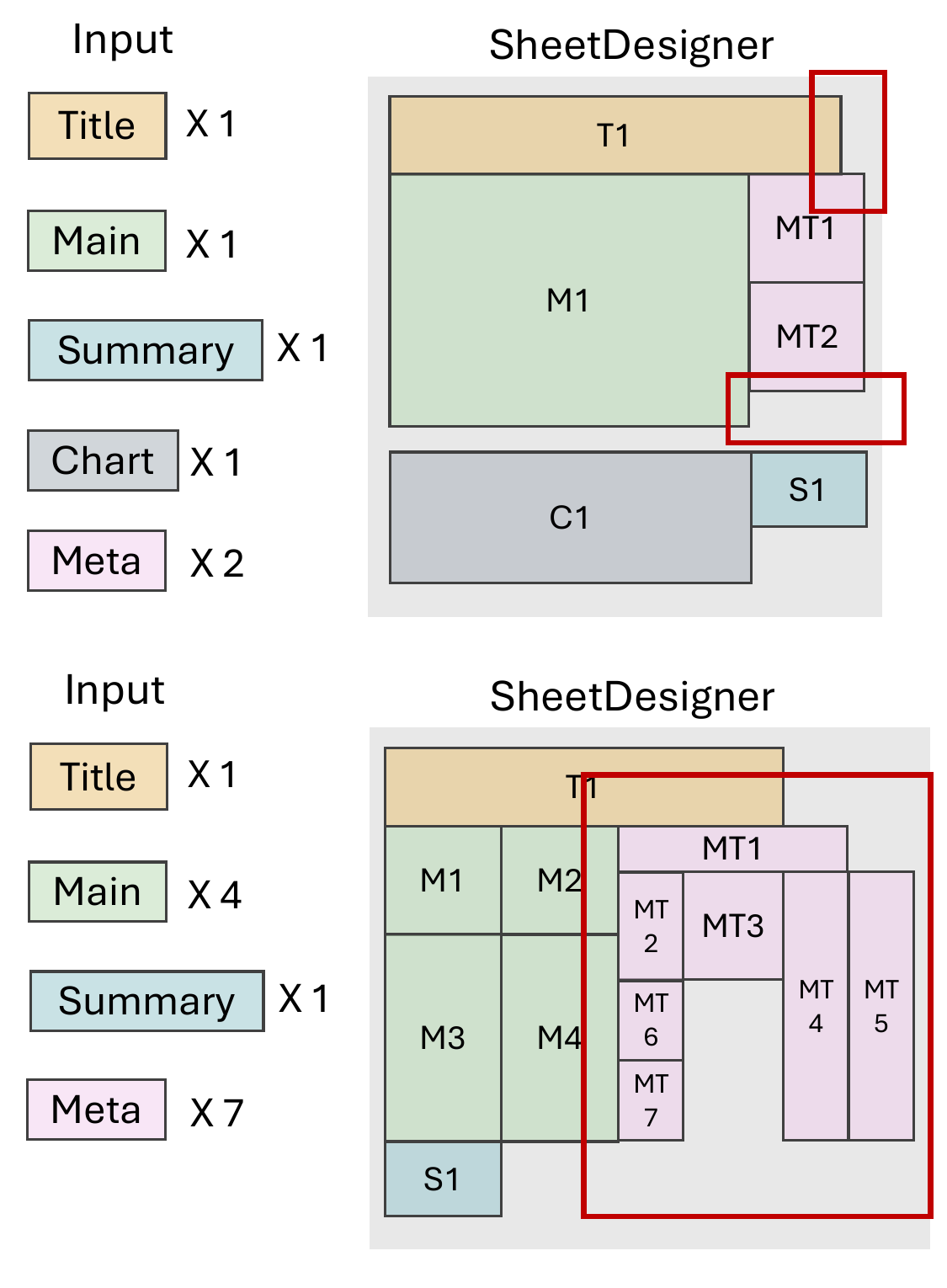}
    \caption{Fail Case study of SheetDesigner}
    \label{fig:fail-case}
\end{figure}

Parallel to the general cases revealed in \autoref{fig:case_good}, in this sub-section we analyze the failure cases of SheetDesigner (with GPT-4o). We find that there are two frequent issues in instances with low scores: (1) some easy-to-get alignment score is not achieved; (2) in cases of extremely enormous components with different sizes, these elements are arranged non-optimally. We provide two examples in \autoref{fig:fail-case}. In the upper figure, there are some easy steps like expanding the title \texttt{T1} to the right border of \texttt{MT1} (Note that title components is allowed to resize), and moving \texttt{MT2} down to align the downsize border of \texttt{M1} will increase the score of alignment. In the lower figure with seven meta-data tables of varying shapes, it fails to provide proper arrangement, leading to a somewhat messy layout, whereas still of certain organization. This analysis reveals the future objectives for improving the SheetDesigner.  

\section{Details of Ablation Study}
\label{sec:appendix_additional_experiments}

Comprehensive details of the ablation study are presented in \autoref{table:ablation-5percent-full} and \autoref{table:ablation-1percent-full}, which report the ablation scores corresponding to the lower 5\% and 1\% tails of the score distributions, respectively.

\begin{table*}[htbp]
    \small
    \centering
    \caption{Ablation study on 5\%-low scores.}
    \label{table:ablation-5percent-full}
    \resizebox{.999\linewidth}{!}{
    \setlength{\tabcolsep}{3pt} 
    \begin{tabular}{lcccccccccccccccccc}
        \toprule
         &  \multirow{2}{*}{\textbf{Fullness}} & \multicolumn{2}{c}{\textbf{Compatibility}} & \multicolumn{2}{c}{\textbf{C-Alignment}} & \multicolumn{2}{c}{\textbf{T-Alignment}} & \multicolumn{2}{c}{\textbf{R-Alignment}} & \multicolumn{2}{c}{\textbf{Balance}} & \multirow{2}{*}{\textbf{Overlap}} & \multirow{2}{*}{\makecell[c]{\textbf{Weighted Total}}} \\
         \cmidrule(l{2pt}r{2pt}){3-4} \cmidrule(l{2pt}r{2pt}){5-6}\cmidrule(l{2pt}r{2pt}){7-8} \cmidrule(l{2pt}r{2pt}){9-10}\cmidrule(l{2pt}r{2pt}){11-12}
         &  & Horizontal & Vertical & Horizontal & Vertical & Horizontal & Vertical & Horizontal & Vertical & Horizontal & Vertical &  & \\ 
         \midrule
SheetDesigner & 0.887 & 0.523 & 0.882 & 0.742 & 0.861 & 0.391 & 0.494 & 0.434 & 0.440 & 0.815 & 0.885 & -0.057 & 4.064 \\
\midrule
w/o Topic & 0.869 & 0.510 & 0.863 & 0.727 & 0.844 & 0.369 & 0.467 & 0.411 & 0.408 & 0.798 & 0.855 & -0.056 & 3.939 ($\downarrow$ 3.06\%)\\
w/o Reflection-Rule & 0.776 & 0.518 & 0.885 & 0.685 & 0.780 & 0.377 & 0.380 & 0.361 & 0.372 & 0.734 & 0.852 & -0.068 & 3.680 ($\downarrow$ 9.44\%)\\
w/o Reflection-Vision & 0.783 & 0.525 & 0.866 & 0.724 & 0.830 & 0.394 & 0.466 & 0.421 & 0.432 & 0.741 & 0.860 & -0.098 & 3.814 ($\downarrow$ 6.13\%)\\
w/o Reflection & 0.755 & 0.522 & 0.867 & 0.656 & 0.765 & 0.357 & 0.365 & 0.355 & 0.356 & 0.719 & 0.839 & -0.083 & 3.572 ($\downarrow$ 12.08\%)\\
w/o SheetRanker & 0.785 & 0.460 & 0.865 & 0.724 & 0.812 & 0.403 & 0.367 & 0.425 & 0.413 & 0.760 & 0.829 & -0.068 & 3.746 ($\downarrow$ 7.81\%)\\
w/o Vision & 0.753 & 0.503 & 0.856 & 0.717 & 0.818 & 0.360 & 0.423 & 0.409 & 0.419 & 0.734 & 0.836 & -0.102 & 3.688 ($\downarrow$ 9.23\%)\\

        \bottomrule
        \end{tabular}
        }
\end{table*}

\begin{table*}[htbp]
    \small
    \centering
    \caption{Ablation study on 1\%-low scores.}
    \label{table:ablation-1percent-full}
    \resizebox{.999\linewidth}{!}{
    \setlength{\tabcolsep}{3pt} 
    \begin{tabular}{lcccccccccccccccccc}
        \toprule
         &  \multirow{2}{*}{\textbf{Fullness}} & \multicolumn{2}{c}{\textbf{Compatibility}} & \multicolumn{2}{c}{\textbf{C-Alignment}} & \multicolumn{2}{c}{\textbf{T-Alignment}} & \multicolumn{2}{c}{\textbf{R-Alignment}} & \multicolumn{2}{c}{\textbf{Balance}} & \multirow{2}{*}{\textbf{Overlap}} & \multirow{2}{*}{\makecell[c]{\textbf{Weighted Total}}} \\
         \cmidrule(l{2pt}r{2pt}){3-4} \cmidrule(l{2pt}r{2pt}){5-6}\cmidrule(l{2pt}r{2pt}){7-8} \cmidrule(l{2pt}r{2pt}){9-10}\cmidrule(l{2pt}r{2pt}){11-12}
         &  & Horizontal & Vertical & Horizontal & Vertical & Horizontal & Vertical & Horizontal & Vertical & Horizontal & Vertical &  & \\ 
         \midrule
SheetDesigner & 0.876 & 0.521 & 0.896 & 0.770 & 0.879 & 0.253 & 0.263 & 0.385 & 0.360 & 0.845 & 0.867 & -0.287 & 3.608\\
\midrule
w/o Topic & 0.867 & 0.521 & 0.893 & 0.756 & 0.858 & 0.255 & 0.256 & 0.338 & 0.333 & 0.800 & 0.795 & -0.293 & 3.477 ($\downarrow$ 3.66\%)\\
w/o Reflection-Rule & 0.829 & 0.525 & 0.887 & 0.611 & 0.756 & 0.187 & 0.189 & 0.301 & 0.345 & 0.720 & 0.746 & -0.347 & 3.115 ($\downarrow$ 13.66\%)\\
w/o Reflection-Vision & 0.812 & 0.523 & 0.890 & 0.759 & 0.854 & 0.241 & 0.267 & 0.384 & 0.352 & 0.709 & 0.723 & -0.417 & 3.246 ($\downarrow$ 10.05\%)\\
w/o Reflection & 0.785 & 0.519 & 0.883 & 0.604 & 0.744 & 0.175 & 0.182 & 0.287 & 0.338 & 0.702 & 0.719 & -0.429 & 2.933 ($\downarrow$ 18.73\%)\\
w/o SheetRanker & 0.830 & 0.503 & 0.880 & 0.759 & 0.831 & 0.249 & 0.232 & 0.380 & 0.372 & 0.780 & 0.815 & -0.363 & 3.368 ($\downarrow$ 6.68\%)\\
w/o Vision & 0.833 & 0.516 & 0.893 & 0.811 & 0.848 & 0.224 & 0.182 & 0.396 & 0.390 & 0.693 & 0.712 & -0.445 & 3.220 ($\downarrow$ 10.75\%)\\

        \bottomrule
        \end{tabular}
        }
\end{table*}

\section{Additional Experiments}
In this section we provide some additional experiments.

\subsection{Choice of Exemplar Images}
We provide the ablation on the choice of exemplar images in \autoref{table:ablation-exemplar}. \autoref{table:ablation-100percel-full} demonstrates that topic-guided exemplar retrieval outperforms methods that do not leverage topic exemplars, achieving a 2.71\% performance improvement. We also provide an additional comparison below, introducing a variant that uses purely random exemplars without regard to topic.

\begin{table}[H]
    \centering
    \caption{Ablation on the choice of exemplar images}
    \label{table:ablation-exemplar}
        \begin{tabular}{lccccccc}
            \toprule
             & Score \\
            \midrule
           SheetDesigner (Topic Exemplar) & 4.898 \\
           SheetDesigner (Random Exemplar) & 4.780 \\
           SheetDesigner (No Exemplar) & 4.766 \\
            \bottomrule
        \end{tabular}
\end{table}

This demonstrates that even a simple topic-based selection strategy offers clear benefits. Although more fine-grained semantic retrieval could further improve performance, it primarily enhances content-level alignment rather than layout structure, making it less directly relevant to the layout generation task. Additionally, such approaches are often computationally intensive and tailored to specific tasks. We therefore consider them more appropriate for future work.

\subsection{Traditional Methods with R1C1 Format}

Additionally, we conducted an experiment equipping LayoutPrompter with the R1C1 coordinate system in \autoref{table:vanilla-r1c1}. While this improved its overall performance, a significant gap remained between SheetDesigner and LayoutPrompter, highlighting the effectiveness of our design beyond the form of coordinate system.

\begin{table*}[htbp]
    \small
    \centering
    \caption{Results of traditional methods with R1C1 format.}
    \label{table:vanilla-r1c1}
    \resizebox{.999\linewidth}{!}{
    \begin{tabular}{lcccccccccccccccccc}
        \toprule
         & {\textbf{Fullness}} & {\textbf{Compatibility}} & {\textbf{C-Alignment}} &{\textbf{T-Alignment}} & {\textbf{R-Alignment}} & {\textbf{Balance}} & {\textbf{Overlap}} & {\textbf{Weighted Total}} \\
         \midrule
SheetDesigner &  0.981	& 0.718&	0.782	&0.823	&0.691	&0.907	&-0.003 & 4.898 \\
\midrule
LayoutPrompter & 0.804	&0.510	&0.649	&0.585	&0.703	&0.706	&-0.167&	3.789 \\
LayoutPrompter(R1C1) & 0.812	&0.509	&0.672	&0.604	&0.713	&0.693	&-0.142 &3.861\\

        \bottomrule
        \end{tabular}
        }
\end{table*}

\subsection{Traditional Transformer-based Methods with More Training Data}

Below, we present the results of an experiment conducted under a label-rich setting, using a 60\%-20\%-20\% train-validation-test split. With more labels available, traditional methods such as LayoutFormer++ and Coarse-to-Fine achieve substantial performance gains; however, they still perform significantly below SheetDesigner.

\begin{table*}[htbp]
    \small
    \centering
    \caption{Results of traditional transformer-based methods with 60\%-20\%-20\% train-validation-test split}
    \label{table:vanilla-tfm-more-data}
    \resizebox{.999\linewidth}{!}{
    \begin{tabular}{lcccccccccccccccccc}
        \toprule
         & {\textbf{Fullness}} & {\textbf{Compatibility}} & {\textbf{C-Alignment}} &{\textbf{T-Alignment}} & {\textbf{R-Alignment}} & {\textbf{Balance}} & {\textbf{Overlap}} & {\textbf{Weighted Total}} \\
         \midrule

SheetDesigner &  0.978	&0.721	&0.769	&0.831	&0.689	&0.901	&-0.004	&4.885 \\
\midrule
LayoutPrompter & 0.803	&0.513	&0.653	&0.594	&0.699	&0.712	&-0.158	&3.816 \\
Coarse-to-Fine & 	0.583	&0.462	&0.573	&0.51	&0.583	&0.652	&-0.132	&3.231\\ 
LayourFormer++ & 0.631	&0.464	&0.63	&0.504	&0.621	&0.69	&-0.116	&3.424\\

        \bottomrule
        \end{tabular}
        }
\end{table*}

\subsection{Statistics of Experimental Results}

We provide the standard deviation of some experimental results of \autoref{table:main} in \autoref{table:statistics-or-experiments}.

\begin{table*}[htbp]
    \small
    \centering
    \caption{Statistics of experimental results.}
    \label{table:statistics-or-experiments}
    \resizebox{.999\linewidth}{!}{
    \begin{tabular}{lcccccccccccccccccc}
        \toprule
         & {\textbf{Fullness}} & {\textbf{Compatibility}} & {\textbf{C-Alignment}} &{\textbf{T-Alignment}} & {\textbf{R-Alignment}} & {\textbf{Balance}} & {\textbf{Overlap}} & {\textbf{Weighted Total}} \\
         \midrule

SheetDesigner &  0.978${\tiny \pm 0.03}$	&0.721${\tiny \pm 0.08}$	&0.769${\tiny \pm 0.12}$	&0.831${\tiny \pm 0.13}$	&0.689${\tiny \pm 0.16}$	&0.901${\tiny \pm 0.08}$	&-0.004${\tiny \pm 0.13}$	&4.885 ${\tiny \pm 0.28}$\\

        \bottomrule
        \end{tabular}
        }
\end{table*}

\section{Experimental Environment Details}

We use GPT-4o via the official OpenAI API\footnote{https://openai.com}, while all other models are run locally on a server equipped with an AMD EPYC 7V13 64-Core Processor, 866 GB of RAM, and four NVIDIA A100 GPUs with a total of 320 GB GPU memory. The experiments required approximately 1,200 GPU hours in total. The cost associated with GPT-4o API usage is estimated at approximately \$2,892.

For GPT-4o, we set a maximum token limit of 16,384 per invocation, with top-p set to 0.95 and a temperature of 0.7. Structured output is enabled. For Vicuna and LLaVA models, we follow the hyperparameter settings provided in their official implementations. The threshold values for Dual LoRA are selected based on a balance between performance and average token cost, as detailed below

\begin{table}[h]
    \centering
    \caption{Token cost analysis of SheetDesigner, reporting average token cost per instance, for a single run, and for a full run with three repeats.}
    \label{table:token-cost}
    \setlength{\tabcolsep}{3pt}
    \resizebox{.999\linewidth}{!}{
    \begin{tabular}{lcccccccccccccccccc}
        \toprule
          & Pre-Process & Structure & Revise & Content & Total (Single) & Total\\             
         \midrule
        Vicuna-7B & 278.8 & 1301.2 & 310.2 & 1080.5 & 2970.7 & 7734.1 \\
        Vicuna-13B & 283.1 & 1339.6 & 262.3 & 1154.5 & 3039.5 & 8027.7 \\
        GPT-4o & 280.5 & 1165.7 & 234.2 & 958.4 & 2638.9 & 6887.3 \\
        \bottomrule
        \end{tabular}
        }
\end{table}

We conduct a token cost analysis running SheetDesigner on different models in \autoref{table:token-cost} \footnote{Note that revision is conditionally triggered; runs without revision are marked with a cost of 0 in this procedure}. We calculate the total token costs using the official OpenAI API response for GPT-4o, while the token length for Vicuna/LLaVA models is determined by the specific tokenizer used. Based on current pricing, a single run with GPT-4o costs approximately 0.0029\$, totaling around 0.0719\$ per instance for a complete execution.

\section{Detailed Prompts}

\label{sec:appendix_prompts}
In this section we provide the detailed prompts for the SheetDesigner, where $\{...\}$ denotes the placeholder to fill in the corresponding data. For all the prompts the input data includes the different stage of the developing spreadsheet layout, from raw spreadsheet data to the revised layout to be populated. For the prompt of Dual Reflection, there are additional inputs of the specific instructions which are triggered by rules, the full set of specific instructions are in \autoref{table:dual-reflection-specific-instructions}. 


\begin{tcolorbox}[
  enhanced,
  colback=white,
  colframe=black,
  boxrule=0.8pt,
  arc=4mm,
  top=2mm,
  bottom=2mm,
  left=3mm,
  right=3mm,
  coltitle=black,
  title=\textbf{Prompts for Pre-processing},
  colbacktitle=black!10,
  fontupper=\small
]

\#\# \textbf{Task}  

You will receive a list of spreadsheet components, each accompanied by comments, descriptions, and detailed data.
Your task is to identify pairs of components that have a logical relationship.
For example, if $\texttt{summary\_table\_1}$ summarizes data from $\texttt{main\_table\_1}$, you should extract and present this relationship as:
$\texttt{(main\_table\_1, summary\_table\_1)}$
 
\#\# \textbf{Hints} 

(1) Relationships can be based on dependencies, references, or summarization within the spreadsheet structure.
(2) If \texttt{component\_A} describes, summarizes, or illustrates data derived from \texttt{component\_B}, then \texttt{A} and \texttt{B} are related.
(3) Organize the results in list of lists, where the inner list should be a 2-component list like \texttt{[A, B]}.
 
\#\# \textbf{Spreadsheet Components}

\{...\}
\end{tcolorbox}

\begin{tcolorbox}[
  enhanced,
  colback=white,
  colframe=black,
  boxrule=0.8pt,
  arc=4mm,
  top=2mm,
  bottom=2mm,
  left=3mm,
  right=3mm,
  coltitle=black,
  title=\textbf{Prompts for Structure Placement},
  colbacktitle=black!10,
  fontupper=\small
]

\#\# \textbf{Task}

I will provide you with a spreadsheet skeleton with multiple elements including title, main-table, meta-data, summary-table, and charts in JSON.
The task is to place the elements by setting their position in the spreadsheet in a good structure.

\#\# \textbf{Instructions}

There are some hints to place the elements:

- The location of elements should be provided via the "location" attribute, which should be a list of two strings indicating the left-top and bottom down corner of the element. Example: ["A1", "C3"].

- The elements placed should align with each other. You can also maintain some symmetry.
	- Specially, maintain a type-aware alignment between element groups. For example, the metadata tables should be aligned with each other.
	- Specially, maintain a relation-aware alignment between elements. For example, the chart demonstrating certain main-table should be aligned with that main-table.
    
- Avoid overlapping the elements.

- The spreadsheet is a 2D grid, so don't place the elements wholly horizontally or vertically. Arrange them in a compound manner.

- When placing the elements, you can leave some space as margins between them. But, avoid leaving too much space empty in the whole spreadsheet.

- Place the elements considering the relationship between them, for example, the summary-table should be placed below the main-table.

- You can change the size of the components following these rules:
	- Title: can be arbitrarily resized.
	- Main-table: you can add empty rows (or namely, changing the height of the table) to make it look good. But, the width should be the same as the given width.
	- Meta-data, summary data: not re-sizable.
    
- The title should be placed at the top of the spreadsheet, spanning all active columns where there are components.

- Do not duplicate the components, each type of components should be placed under the corresponding lists.

\#\# \textbf{Spreadsheet Skeleton Set}

\{...\}

\end{tcolorbox}

\begin{tcolorbox}[
  enhanced,
  colback=white,
  colframe=black,
  boxrule=0.8pt,
  arc=4mm,
  top=2mm,
  bottom=2mm,
  left=3mm,
  right=3mm,
  coltitle=black,
  title=\textbf{Prompts for Dual Reflection},
  colbacktitle=black!10,
  fontupper=\small
]
\#\# \textbf{Task}

I will provide you with a spreadsheet layout with multiple elements including title, main-table, meta-data, summary-table, and charts in JSON.
Your task is to revise the structure following the instructions. I will first provide you with the general instructions,
then is the specific instructions I want you to follow. You will need to revise the structure of the spreadsheet accordingly.

\#\# \textbf{General Instructions}

<Instructions for structure placement>

\#\# \textbf{Specific Instructions}

\{...\}

\#\# \textbf{Spreadsheet layout}

\{...\}

\end{tcolorbox}

\begin{tcolorbox}[
  enhanced,
  colback=white,
  colframe=black,
  boxrule=0.8pt,
  arc=4mm,
  top=2mm,
  bottom=2mm,
  left=3mm,
  right=3mm,
  coltitle=black,
  title=\textbf{Prompts for Content Population},
  colbacktitle=black!10,
  fontupper=\small
]
\#\# \textbf{Task}

I will provide you with a spreadsheet with multiple elements including title, main-table, meta-data, summary-table, and charts in JSON.
These components include their location in the spreadsheet, as well as their content.
Your task is to generate the proper line heights and column widths for the spreadsheet, as well as adding line breaks to the content.
The fundamental goal is to (1) make the cells compatible with the content, and (2) make the spreadsheet visually appealing.

\#\# \textbf{Instructions}

There are some hints:
- Carefully consider the content of each cell and adjust the row height and column width accordingly.
- For a cell with lengthy content, you can either wrap the text for a line break, or increase the column width.
- Try assigning line heights and column widths and line breaks to make (1) the spreadsheet is balanced vertically and horizontally, and (2) the neither too compact nor too much empty spaces..
- You may assume a default font setting of Calibri 11 and excel standard column width and row heights, where 1 character
of text are compatible with 0.65 of column width.

\#\# \textbf{Spreadsheet Skeleton Set with Contents}

\{...\}

\end{tcolorbox}

\section{Future Works}
Future work can extend this research in several promising directions:
\begin{itemize}
    \item Graph-based Representation Learning: We plan to leverage the heterogeneous graph structure of spreadsheet components \cite{chen2025adaptive} and apply advanced graph learning techniques \cite{DBLP:conf/nips/LiuCYY24,chen2025dagprompt}. This will enable a more comprehensive modeling of component relationships and facilitate more powerful representation learning.
    \item Expanded Ethical Analysis: We aim to broaden the ethical analysis to explore other critical dimensions of AI safety, such as the problem of value and preference alignment between humans and AI systems \cite{ren-etal-2024-valuebench}.
\end{itemize}

\end{document}